\documentclass[journal]{IEEEtran}
\usepackage{times}
\usepackage{epsfig}
\usepackage{graphicx}
\usepackage{amsmath}
\usepackage{amssymb}
\usepackage{subfigure}
\usepackage{cite}
\usepackage{multirow}
\usepackage{color}
\usepackage{xcolor}
\usepackage{soul}
\usepackage{algorithm}
\usepackage{algorithmic}
\usepackage{caption}
\usepackage{mathtools}
\usepackage{setspace}
\usepackage{enumitem}
\usepackage{booktabs}
\usepackage[pagebackref=true,breaklinks=true,letterpaper=true,colorlinks,bookmarks=false]{hyperref}
\newcommand{\eg}{\textit{e}.\textit{g}.}

\ifCLASSINFOpdf
\else
\fi

\begin{document}
\title{CERL: A Unified Optimization Framework for Light Enhancement With Realistic Noise}

\author{Zeyuan~Chen,
        Yifan~Jiang,
        Dong~Liu, \IEEEmembership{Senior Member, IEEE,}
        and Zhangyang~Wang, \IEEEmembership{Senior Member, IEEE}
\thanks{Z. Chen is with the School
of the Gifted Young, University of Science and Technology of China.
(e-mail: nnice1216@mail.ustc.edu.cn)}
\thanks{Y. Jiang and Z. Wang are with the Department of Electrical and Computer Engineering, the University of Texas at Austin, TX, USA.
(e-mail: \{yifanjiang97,atlaswang\}@utexas.edu)}
\thanks{D. Liu is with the Department of Electronic Engineering and Information Science, University of Science and Technology of China.
(e-mail: dongeliu@ustc.edu.cn)}
\thanks{Correspondence addressed to D. Liu and Z. Wang.}}

\markboth{IEEE Transactions on Image Processing}%
{Chen \MakeLowercase{\textit{et al.}}: CERL: A Unified Optimization Framework for Light Enhancement With Realistic Noise}

\maketitle

\begin{abstract}
Low-light images captured in the real world are inevitably corrupted by sensor noise. Such noise is spatially variant and highly dependent on the underlying pixel intensity, deviating from the oversimplified assumptions in conventional denoising. Existing light enhancement methods either overlook the important impact of real-world noise during enhancement, or treat noise removal as a separate pre- or post-processing step. We present \underline{C}oordinated \underline{E}nhancement for \underline{R}eal-world \underline{L}ow-light Noisy Images (CERL), that seamlessly integrates light enhancement and noise suppression parts into a unified and physics-grounded optimization framework. For the real low-light noise removal part, we customize a self-supervised denoising model that can easily be adapted without referring to clean ground-truth images. For the light enhancement part, we also improve the design of a state-of-the-art backbone. The two parts are then joint formulated into one principled plug-and-play optimization. Our approach is compared against state-of-the-art low-light enhancement methods both qualitatively and quantitatively. Besides standard benchmarks, we further collect and test on a new realistic low-light mobile photography dataset (RLMP), whose mobile-captured photos display heavier realistic noise than those taken by high-quality cameras. CERL consistently produces the most visually pleasing and artifact-free results across all experiments. Our RLMP dataset and codes are available at: \url{https://github.com/VITA-Group/CERL}.
\end{abstract}

\ifCLASSOPTIONpeerreview
\begin{center} \bfseries EDICS Category: 3-BBND \end{center}
\fi

\IEEEpeerreviewmaketitle

\section{Introduction}
\IEEEPARstart{L}{ow-light} images are generally degraded by low contrast and poor visibility, causing unpleasant subjective feelings of people. In past decades, many methods have been proposed to solve the problem of low-light image enhancement. Classic methods \cite{pizer1987adaptive}, \cite{jobson1997properties}, \cite{jobson1997multiscale} directly increase the brightness and contrast of low-light images, but they are not robust across all circumstances. Recently, the advances in deep learning motivate researchers to propose learning-based approaches for low-light image enhancement. A series of methods \cite{lore2017llnet}, \cite{wei2018deep}, \cite{cai2018learning}, \cite{ren2019low}, \cite{wang2019underexposed} learn to enhance images based on paired high/low-light data. They produce more realistic enhancement results with better efficiency compared with the classic methods.

\begin{figure*}
\centering
\includegraphics[width=1.0\linewidth]{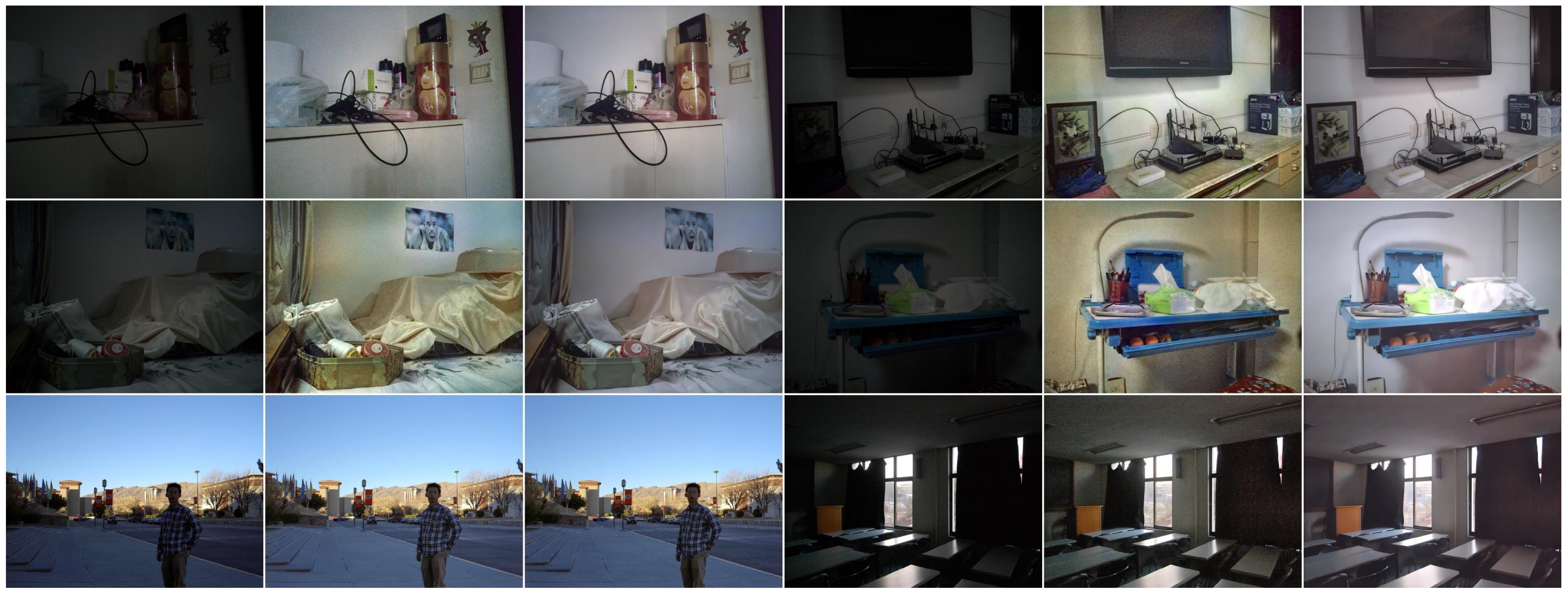}
\caption{Representative visual examples by enhancing low-light noisy images using CERL. From left to right: columns 1,4 are the low-light noisy input images; columns 2,5 are corresponding enhanced results by EnlightenGAN~\cite{jiang2021enlightengan}; and columns 3,6 are the results by CERL. CERL well suppresses the low-light noise while restoring the brightness and colors (\textbf{best zoom-in}).}
\label{fig1}
\end{figure*}

One long-existing problem in the light enhancement task is that real low-light images are inevitably corrupted by sensor noise since there is insufficient light reaching camera sensors in low-light conditions, causing the scene signals buried by heavy noise. The aforementioned light-enhancement methods based on paired training usually fail to handle low-light images with heavy sensor noise, as their synthetic training data is not degraded by real low-light noise. Although we could add simulated noise to the training images, a large domain gap exists between synthetic and real-world noise.

Recently, \cite{wei2018deep} collected a real-world paired dataset for low-light enhancement, named LOL, providing a possible solution for enhancement models to learn denoising by supervised training. However, models trained on a specific dataset typically fail to handle noise from other domains, and poorly generalize its denoising ability to other data. In addition, two categories of methods are proposed to solve the problem of low-light noise. Methods in the first category \cite{guo2016lime,wei2018deep} treat the denoising part as a post-processing step. A common drawback for them is that most existing denoisers fail to well tackle low-light noise which is typically spatially varying. Besides, this two-stage heuristic can lead to sub-optimal results. Methods of the second category \cite{jiang2021enlightengan,guo2020zero,yang2020fidelity} train light-enhancement models without paired supervision. Therefore, their training data can be the unpaired real low-light noisy image and high-light clean image, which helps models learn to adapt to the real low-light noisy image. However, these methods do not explicitly address noise removal as an individual task, hence the denoising performance may be limited.

In this paper, we propose \textit{Coordinated Enhancement with Real Low-light noise} (CERL), that coordinates and integrates low-light enhancement and noise suppression into a unified optimization framework. Specifically, we decouple  light enhancement and noise removal as two sub-problems that will be alternatively handled. To solve them, we present a self-supervised training scheme based on the Retinex theory \cite{land1977retinex}. It fine-tunes a pre-trained denoiser to make it easily adapted to the low-light noise domain. We also modify a state-of-the-art light enhancement backbone for more realistic and artifact-free results. With the two cornerstones of a self-tuned denoiser and a modified light enhancement backbone, we jointly formulate the two sub-problems into a principled plug-and-play optimization problem that can be iteratively solved. By leveraging this optimization framework, we demonstrate that the two highly-entangled elements in real-world light enhancement, noise and light, can be divided and addressed separately. Also, we show that the plug-and-play framework is eminently suitable for this task, as it seamlessly integrates off-the-shelf deep neural networks into the optimization process for optimal enhancement results.

Moreover, existing light enhancement benchmarks are typically collected with high-end cameras equipped with long and stable exposure.
Meanwhile, smart phones have now become the primary source of daily photos. To account for this domain gap, we collect a new \textit{realistic low-light mobile photography} (RLMP) dataset, where photos are captured by multiple common mobile phones. We find our collected photos to display much more noticeable ISO noise, that complements the existing benchmarks and significantly challenges current enhancement methods. The first two rows in Fig.~\ref{fig1} present visual examples from RLMP, and their enhanced results by several methods. 

We summarize the contributions of our work as follows.
\begin{itemize}
    \item \textbf{Framework:} We propose \underline{C}oordinated \underline{E}nhancement for \underline{R}eal-world \underline{L}ow-light Noisy Images (CERL), which explicitly formulates the real-world noise suppression and the light enhancement as two individual problems. CERL unifies the two sub-problems under one principled optimization form, which leads to an iterative plug-and-play framework. By leveraging this framework, we decouple the highly-entangled two elements of noise and light, and achieve their corresponding enhancement objectives iteratively with customized deep neural networks.
    
    \item \textbf{Technique:} Under the unified framework, we present techniques for solving both sub-problems: a self-supervised deep low-light denoiser inspired by the Retinex theory, that can be adapted via self-supervision; as well as an improved light enhancement backbone. The fine-tuned denoiser not only perform well on low-light noise removal, but alleviate the color shifting and distortion problem in the light enhancement backbone.
    
    \item \textbf{Performance:} Our approach is compared favorably against state-of-the-art low-light enhancement methods, both qualitatively and quantitatively. At the presence of low-light noise, it consistently produces the most visually pleasing and artifact-free results, on both existing and our newly collected RLMP dataset. Besides, CERL gains solid performance gains in both the light and noise parts of this task compared with the selected light enhancement and denoising backbones. 
\end{itemize}

\begin{figure*}
\centering
\includegraphics[width=0.97\linewidth]{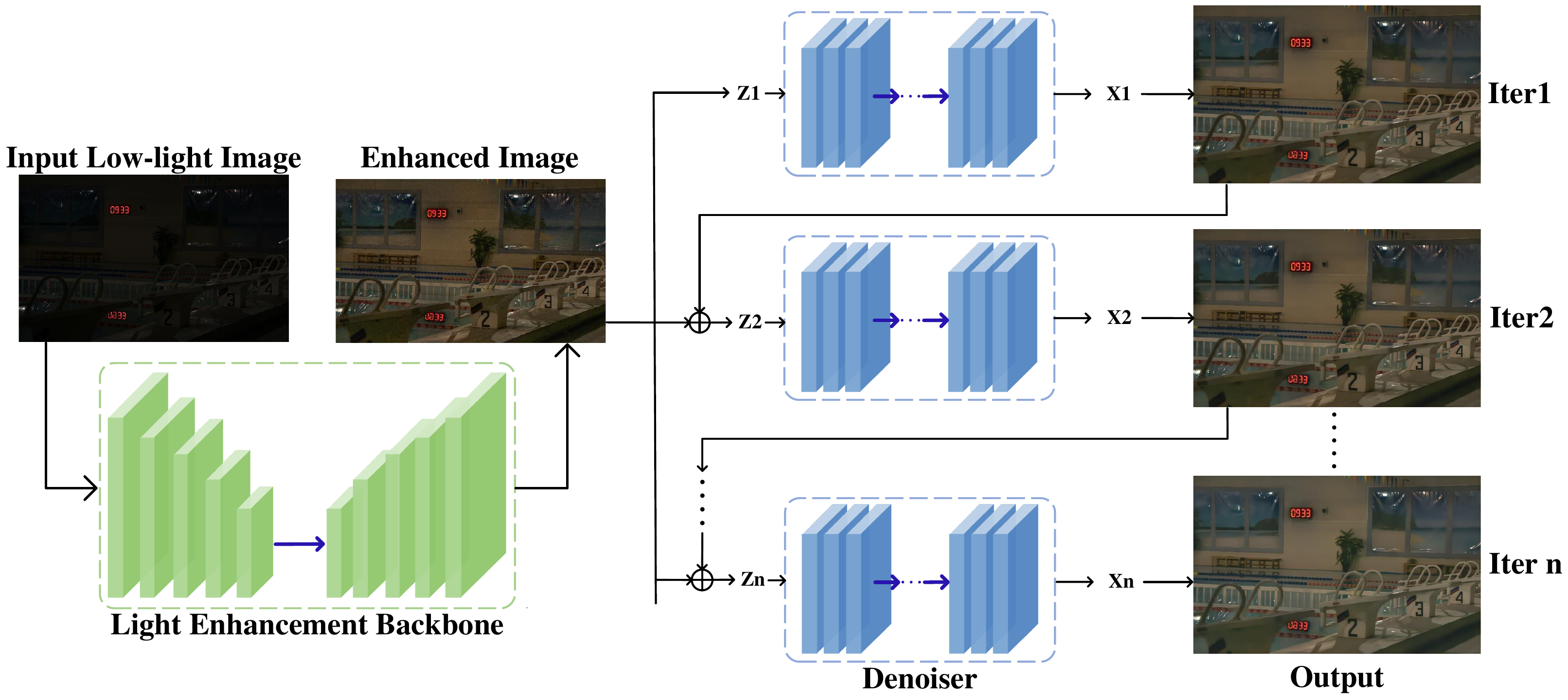}
\caption{Overview of the proposed CERL framework. In this paper, EnlightenGAN \cite{jiang2021enlightengan} is adapted and used as the light enhancement backbone, and the denoising model in \cite{zhou2020awgn} is fine-tuned as the denoiser. Note that the step of (re-)adding noise is not shown explicitly in the figure.}
\label{fig_framework}
\end{figure*}

\section{Related Work}
\textbf{Traditional Low-Light Enhancement:} The low-light image enhancement task is studied as an image processing problem for a long while. Histogram equalization (HE) based methods \cite{pizer1987adaptive,stark2000adaptive,abdullah2007dynamic} perform light enhancement by expanding the dynamic range of an image. Retinex theory-based methods \cite{jobson1997multiscale, jobson1997properties} decompose an image into illumination and reflectance layers, and adjust them respectively to obtain an enhanced high-light image. \cite{wang2013naturalness} proposed an enhancement algorithm for non-uniform illumination images, utilizing a bi-log transformation to map the illumination and make a balance between details and naturalness. LIME \cite{guo2016lime} is an effective low-light enhancement method, where the illumination map is initialized and then refined by imposing a structure prior. \cite{li2018structure} proposed a robust Retinex model, which additionally considers a noise map to improve the model performance on low-light images accompanied by intensive noise.

\textbf{Learning-Based Low-Light Enhancement:} 
With the availability of large-scale paired data and powerful CNNs, learning-based methods have become popular in recent years~\cite{ouyang2018pedestrian}, \cite{jiang2021transgan}, \cite{jiang2021ssh}, \cite{jiang2022fast}, \cite{kupyn2019deblurgan}, \cite{Li_2017_ICCV}, \cite{chen2021psd}. LL-Net \cite{lore2017llnet} is a deep autoencoder-based network that jointly learns denoising and light enhancement on the patch level.  \cite{wei2018deep} proposed an end-to-end Retinex-Net that incorporates the Retinex model into the network architecture. Other methods \cite{cai2018learning}, \cite{lv2018mbllen}, \cite{ren2019low}, \cite{wang2019underexposed} based on various network designs are also presented to solve this task. \cite{chen2018learning} developed a pipeline for processing low-light raw images. Recently, several enhancement models without paired supervision are proposed. EnlightenGAN \cite{jiang2021enlightengan} is the first attempt that uses unpaired data to train a low-light enhancement model, of which the network architecture is based on the generative adversarial network. \cite{guo2020zero} presented Zero-DCE, formulating the light enhancement as a task of image-specific curve estimation with a deep network.  \cite{yang2020fidelity} designed a semi-supervised learning framework called DRBN, which extracts a series of coarse-to-fine band representations for the enhancement. As the above methods show impressive performance on light enhancement, they mostly fail to handle low-light noise well when increasing the image brightness. We aim at producing clean and artifact-free enhancement results on low-light images with real-world noise, and we experimentally compare with some state-of-the-art methods.

\textbf{Low-Light Noise Removal: }
Low-light images are inevitably degraded by high ISO noise that differs physically from the conventional simulated noide models \cite{liu2018image,liu2020connecting}. Approaches that focus on generic real-world denoising~\cite{lehtinen2018noise2noise}, \cite{cai2021learning}, \cite{jia2021ddunet}, \cite{jia2021pixel} can be adopted to address this issue, but their performances are typically limited due to the lack of domain knowledge in low-light noise. Previous low-light enhancement methods \cite{guo2016lime,wei2018deep} apply the popular denoising method BM3D \cite{dabov2007image} on the enhanced images. LL-Net \cite{lore2017llnet} consists of two sub-networks dedicated to light enhancement and noise removal, respectively. These two-stage methods usually produce sub-optimal enhancement results. \cite{wei2018deep} collected the LOL dataset containing paired low/normal-light images taken from real scenes. With the support of the LOL dataset, some deep networks \cite{yang2020fidelity,zhang2019kindling} succeed in suppressing the noise in the enhancing results. KinD \cite{zhang2019kindling} learns to remove low-light noise by a specially designed smoothness loss function. \cite{wang2019progressive} presented a progressive retinex model for noise removal, but their network is trained on synthetic paired data, hence its performance is limited when applied to real-world low-light images. DRBN \cite{yang2020fidelity} is trained to suppress the image noise by the residual learning scheme.  However, they typically fail to generalize to other noise distributions beyond the LOL dataset. Different from the above methods, We propose a self-supervised fine-tuning scheme for denoising, intending to obtain a deep low-light denoiser that generalizes well on low-light images with real noise. Also, the presented CERL framework helps to prevent from generating sub-optimal enhancement results.

\section{Method}
In this section, we start by analyzing the degradation model of real-world low-light enhancement problem, and then introduce the proposed CERL framework as a solver. A self-supervised denoiser and a modified light enhancer are followed to construct the proposed framework. The overview of CERL is illustrated in Fig.~\ref{fig_framework}.

\subsection{Coordinated Optimization Framework}
Low-light images are inevitably degraded by noise. Since there is usually insufficient light coming into the camera sensors, noticeable system noise is unavoidable when captured in low-light environments. Different from the usual additive white Gaussian noise (AWGN) model, real-world noise in low-light images is much more sophisticated, as it is spatially variant and signal-dependent. Moreover, when enhancing a low-light image, the original noise can also be amplified, suggesting that light enhancement and noise removal are two mutually entangled sub-problems in the real-world low-light enhancement task. 
From the above perspective, we express the degradation model at low light conditions with noise in the following generic way:
\begin{equation}
    \boldsymbol{y} = \mathcal{L}(n(\boldsymbol{x}))
\label{eq0}
\end{equation}
where $\boldsymbol{x}$ and $\boldsymbol{y}$ represent high-light clean image and low-light noisy image, respectively. $\mathcal{L}(\cdot)$ represents the light reduction and $n(\cdot)$ is the function that adds noise to an image.

A common solution for Eq. (\ref{eq0}) is to first perform the light enhancement and subsequently apply an off-the-shelf denoiser as a post-processing method. However, this naive strategy often leads to sub-optimal results due to the two sub-problems are high-dimensional entangled. 
In order to decouple the two problems, we formulate the energy function according to the maximal a posteriori (MAP) criterion
\begin{equation}
    E(\boldsymbol{x}, \boldsymbol{y}) = \frac{1}{2}\left\|\boldsymbol{y}-\mathcal{L}(n(\boldsymbol{x}))\right\|^2 +\lambda\Phi(\boldsymbol{x}) 
\label{eq_energy}
\end{equation}
where $\frac{1}{2}\left\|\boldsymbol{y}-\mathcal{L}(n(\boldsymbol{x}))\right\|$ is the data fidelity term determined by Eq. (\ref{eq0}) and $\Phi(x)$ is the prior term. $\lambda$ is the regularization parameter. 

\begin{figure*}[h]
\centering
\includegraphics[width=1.0\linewidth]{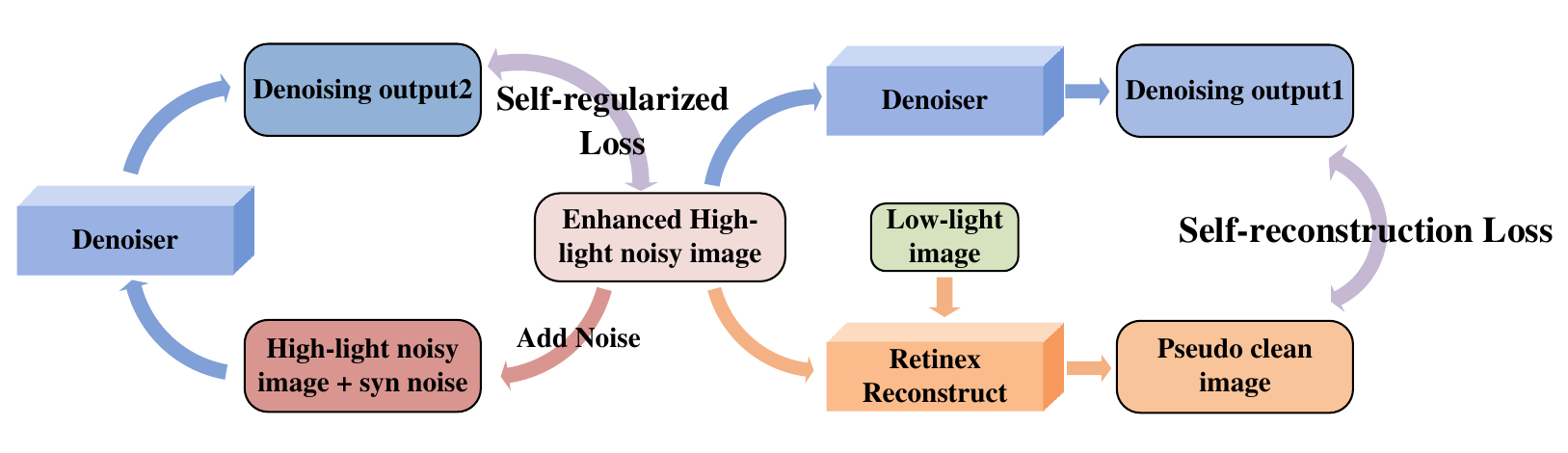}
\caption{The self-supervised fine-tuning scheme for our denoiser.}
\hspace{-0.5cm}
\label{fig_method}
\end{figure*}

To minimize Eq.~(\ref{eq_energy}), we adopt the variable splitting technique and introduce an auxiliary variable $\boldsymbol{z}$, resulting in the following optimization formulation
\begin{align}
    \min_{\boldsymbol{x}}\frac{1}{2}\left\|\boldsymbol{y}-\mathcal{L}(\boldsymbol{z})\right\|^2+\lambda\Phi(\boldsymbol{x}) \nonumber\\
    \mbox{subject to } \boldsymbol{z} = n(\boldsymbol{x})
\label{eq3}
\end{align}
We then use the half quadratic splitting (HQS) algorithm to address Eq. (\ref{eq3}), by minimizing the following function
\begin{equation}
    L_{\mu}(\boldsymbol{x},\boldsymbol{z})=\frac{1}{2}\left\|\boldsymbol{y}-\mathcal{L}(\boldsymbol{z})\right\|^2+\lambda\Phi(\boldsymbol{x})+\frac{\mu}{2}\left\|\boldsymbol{z}-n(\boldsymbol{x})\right\|^2
\label{eq4}
\end{equation}
where $\mu$ is the penalty parameter. In practice, $\mu$ is set to a non-descending series during the iterations to ensure model convergence. The iterative solution is
\begin{align}
    \label{iter1}\boldsymbol{z}_{k+1}&=\arg\min_{\boldsymbol{z}}\frac{1}{2}\left\|\boldsymbol{y}-\mathcal{L}(\boldsymbol{z})\right\|^2+\frac{\mu}{2}\left\|\boldsymbol{z}-n(\boldsymbol{x}_k)\right\|^2  \\
    \label{iter2}\boldsymbol{x}_{k+1}&=\arg\min_{\boldsymbol{x}}\frac{\mu}{2}\left\|\boldsymbol{z}_{k+1}-n(\boldsymbol{x})\right\|^2 +\lambda \Phi(\boldsymbol{x})
\end{align}
where $\boldsymbol{x}_{k}$ is the recovered high-light clean image $\boldsymbol{x}$ at iteration $\textit{k}$. $\boldsymbol{z}_{k}$ is the value of auxiliary variable $\boldsymbol{z}$ at iteration $\textit{k}$.

Eqs.~(\ref{iter1}) and (\ref{iter2}) are alternating minimization problems with respect to $\boldsymbol{z}$ and $\boldsymbol{x}$. For Eq.~(\ref{iter1}), with the assumption that $\mathcal{L}$ is invertible, one can find that the inverse function of $\mathcal{L}$ corresponds to a light enhancing function, which could be substituted by a light enhancement model $E(\cdot)$. If this is the case, we may write a closed-form solution for Eq.~(\ref{iter1}) as
\begin{equation}
    \boldsymbol{z}_{k+1}=\frac{E(\boldsymbol{y})+\mu n(\boldsymbol{x}_{k})}{1+\mu}
\label{eq5}
\end{equation}
where $E(\boldsymbol{y})$ is the enhanced result of the corresponding low-light image and $n(\cdot)$ represents the function that adds noise to an image\footnote{Theoretically the real-world noisy data is needed, here we add spatially variant Gaussian noise instead. See Section \ref{sec31} for details.}. We can observe that the solution of this problem only depends on the solver for the light enhancement problem and is unrelated to the performance of denoising model. 

For the analysis of Eq. (\ref{iter2}), one could clearly see that it corresponds to a denoising problem: we have a noisy image $\boldsymbol{z}$ and aim to restore the latent clean image $\boldsymbol{x}$. Therefore, we can plug-in any well-trained deep denoiser to solve the problem, and rewrite the equation as 
\begin{equation}
    \boldsymbol{x}_{k+1}=\mbox{Denoiser}(\boldsymbol{z}_{k+1})
    \label{eq6}
\end{equation}
This solution corresponds to a pure denoising process and does not depend on the solution of light enhancement.

To this end, we show that the two sub-problems --- light enhancement and noise removal --- are decoupled through a \textbf{coordinated} iterative optimization framework according to Eqs.~(\ref{eq5}) and (\ref{eq6}). Our entire framework, as illustrated in Fig.~\ref{fig_framework}, is aligned with the popular idea of plug-and-play optimization in physics-grounded computational imaging \cite{zhang2019deep,ryu2019plug}

\begin{figure*}[h]
\centering
\includegraphics[width=1\linewidth]{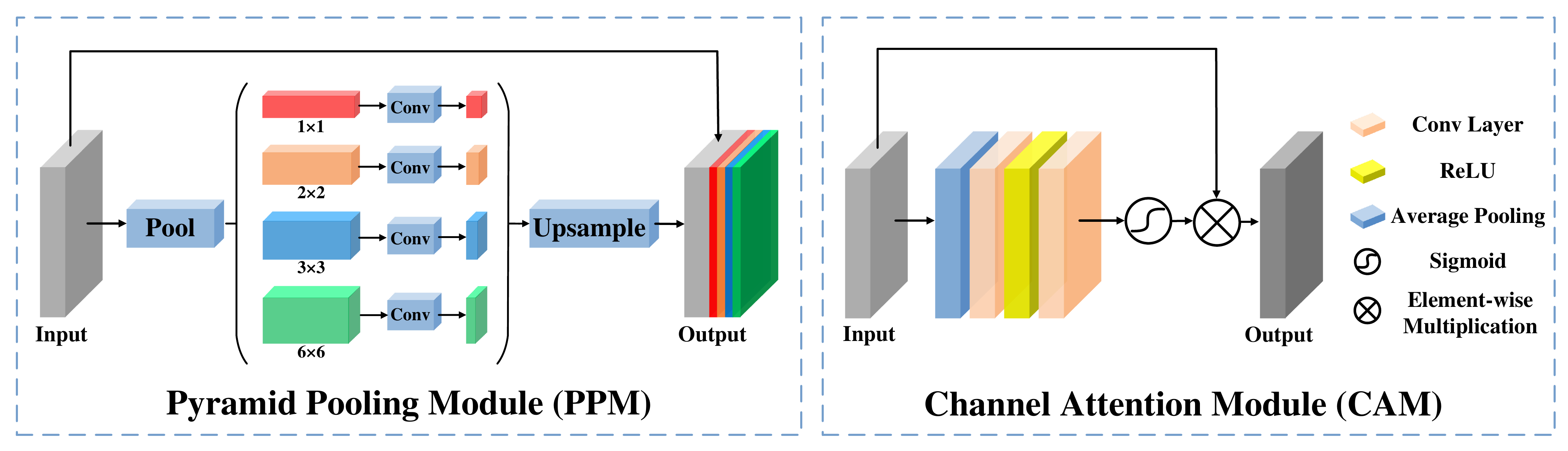}
\caption{The architectures of pyramid pooling module and channel attention module that are employed by our proposed EnlightenGAN+.}
\label{net}
\end{figure*}

\subsection{Self-supervised Denoiser Fine-tuning}\label{sec31}
Although a substantial amount of models are proposed to handle images corrupted by various types of noise, few of them is dedicated to removing low-light noise in real world, which is often hard to tackle especially when the noise is amplified by the light enhancement process. In order to address the problem, we take a denoising model pre-trained on synthetic white Gaussian noise, and fine-tune the model in a self-supervised fashion, adapting it to the low-light noise domain.   
An overview of the self-supervised fine-tuning scheme is presented in Fig.~\ref{fig_method}. We adopt two specially designed loss functions to guide the self-supervised fine-tuning.\\
\textbf{Self-reconstruction Loss.}
We reconstruct pseudo high-light clean images based on the Retinex theory, and make them the substitution of the unseen ground truth during the self-supervised learning process.  

To be specific, the Retinex theory decomposes an image into two separated layers
\begin{equation}
    S = R \circ I
\label{eq}
\end{equation}
where $S$ is the source image. $R$ represents reflectance, $I$ represents illumination, and $\circ$ is the element-wise multiplication. The decomposing process could be accomplished by a neural network~\cite{wei2018deep}.

For an image triplet containing a low-light clean image $S_l$, a high-light clean image $S_h$, and an enhanced high-light noisy image $S_e$, we have the following two observations:
\begin{itemize}
    \item Low-light clean image $S_l$ and high-light clean image $S_h$ should share one common reflectance map, meaning that we have $R_l=R_h$.
    \item High-light clean image $S_h$ and enhanced high-light noisy image $S_e$ are supposed to have similar illumination maps, that is, $I_h=I_e$.
\end{itemize}
Based on the above observations, we have
\begin{equation}
    S_h=R_h\circ I_h \approx R_l \circ I_e
\label{eq2}
\end{equation}
From Eq.~(\ref{eq2}), we observe that high-light image $S_h$ could be approximated by $R_l$ and $I_e$, which is available given $S_l$ and $S_e$. Therefore, we can reconstruct the pseudo high-light clean image by $S_{pseudo}=R_l\circ I_e$. As we hope the denoising result of enhanced image $S_e$ to be close to the high-light clean image, the self-reconstruction loss is formulated as
\begin{equation}
    \mathcal{L}_{recon} = \left\| \mathcal{D}(S_{e}) - S_{pseudo}\right\|^2_2
\end{equation}
where $\mathcal{D}$ is the denoiser and $\mathcal{D}(S_{e})$ is the denoising result of $S_e$.
This self-reconstruction loss helps our denoiser to obtain information from the target high-light clean image domain, as well as ensures the denoising results are physically correct based on the Retinex theory.

\textbf{Self-regularized Loss.} Although self-reconstruction loss provides supervision for the unlabeled noisy input, the reconstructed pseudo clean image $S_{pseudo}$ can not be reliable in all circumstances, as the Retinex theory is not very robust. Hence, we impose a regularization during the fine-tuning, aiming at maintaining the whole process more robust. To be more specific, the enhanced result corrupted by low-light noise is directly taken as the ''clean'' target, while the input for our denoiser is the synthetic image consisted of this corrupted enhanced image and a second and similar corruption. We add simulated spatially variant Gaussian noise on the enhanced noisy image to synthesize the desired image input that is corrupted twice and formulate the self-regularized loss by minimizing the distance between the denoiser output and the original corrupted enhancing result
\begin{equation}
    \mathcal{L}_{reg} = \left\| \mathcal{D}(S_{ne}) - S_{e}\right\|^2_2
\end{equation}
Where $S_e$ is the enhanced high-light noisy image, $S_{ne}$ is the input image of denoiser that is corrupted by both real low-light noise and simulated noise. $\mathcal{D}$ is the denoiser to be fine-tuned and $\mathcal{D}(S_{ne})$ is the denoising result of $S_{ne}$. 

The intuition of using such self-regularized loss is based on the observation that when the noise is weak, it is often feasible to learn a well-performed self-supervised denoiser, of which the parameters are approximate to the optimal parameters learned with supervision. Further demonstrations have been presented in \cite{xu2020noisy}. Therefore, the loss function itself could boost the performance of denoiser on low-light noise, as such noise is typically of weak intensity. In addition, this loss works as a regularization term for the training, as it forces the denoiser to generate denoised results close to the enhanced noisy images, which are finally applied to be the input of denoiser during the testing stage. Therefore, the final denoised results are under the control of the original inputs.

Finally, the total loss can be written as
\begin{equation}
    \mathcal{L} = \lambda\mathcal{L}_{recon} + \mathcal{L}_{reg}
\end{equation}
where $\lambda$ is the coefficient. Empirically, it is set to 0.3.

\subsection{EnlightenGAN+: Light Enhancement Backbone}\label{33}

In CERL, except the denoiser, we also need a solver for the light enhancement task, namely a light enhancement backbone. In order to detach from paired supervision, we favor the EnlightenGAN \cite{jiang2021enlightengan} as the backbone model. However, the results of EnlightenGAN still suffer from visible artifacts in many cases. To reduce these artifacts, we make several modifications on the original model and obtain an improved new network named EnlightenGAN+. The modifications are presented below.

\textbf{Pyramid Pooling Module.}
The performance of EnlightenGAN is sometimes degraded by local artifacts. One effective way to eliminate them is to integrate non-local information into the deep network. For this purpose, we incorporate the feature pyramid pooling module \cite{zhao2017pyramid,kupyn2019deblurgan} to help the model capture local context details as well as global structures, thus avoiding unrealistic enhancement results. The module architecture is shown in Fig.~\ref{net}. Specifically, it obtains features under different scales by applying convolution operators of different kernel size, and then fuses these features to capture both local and non-local information of the input image. The module is placed at the bottleneck of our network.

\textbf{Channel Attention.}
Features of different channels contain information with different levels of importance for the model performance. Taking a color image as an example, the channel with the highest intensity usually contains more information about the image illumination condition compared with the other two channels. Consequently, we use a channel attention module \cite{woo2018cbam} to provide channel-wise global information for the network. 

We present the module architecture in Fig.~\ref{net}. In detail, the average pooling operation is adopted to aggregate channel-wise spatial information into a channel descriptor. For an input feature map $F^{in}$ with shape $C \times H \times W$, we have
\begin{equation}
    g_c = H_p(F_c^{in}) = \frac{1}{H \times W} \sum\limits_{i=1}^H \sum\limits_{j=1}^W X_c(i,j)
\end{equation}
where $X_c(i,j)$ is the value of $c$-th channel $X_c$ at position $(i,j)$. $H_p$ is the global pooling function. After the pooling operation, we have a new feature map $g_c$ with shape $C \times 1 \times 1$. 

We then pass $g_c$ through two convolutional layers plus Sigmoid and ReLu activation functions, in order to get the weights for different channels of the input feature maps
\begin{equation}
    W_c = \sigma(Conv(\delta(Conv(g_c))))
\end{equation}
where $\sigma$ is the Sigmoid funtion and $\delta$ is the ReLu function. $W_c$ represents the weight for the $c$-th channel.

Finally, we element-wise multiply the input and the channel weights to obtain each channel $F_c^{out}$ of the output feature maps $F^{out}$
\begin{equation}
    F_c^{out} = W_c \otimes F_c^{in}
\end{equation}

\textbf{Bright Channel Loss.}
Bright channel prior is an effective prior for low-light enhancement. It supposes that for a well-illuminated image, the maximum intensity in RGB channels of each pixel should be close to the maximal allowed value. It is widely applied to solve the low-light enhancement task in conventional methods \cite{wang2013automatic,tao2017low}. We adopt the prior as a loss function to guide the training of the enhancement network. This bright channel loss is supposed to help the network better restore illumination information and avoid under-exposure problems that sometimes exists in the results of vanilla EnlightenGAN. The loss is formulated as
\begin{equation}
    \mathcal{L}_{bright}(I) = \sum\limits_{\boldsymbol{x}\in I}1 - \max\limits_c\max\limits_{\boldsymbol{y}\in \Omega(\boldsymbol{x})}(I_c(\boldsymbol{y}))
\end{equation}
where $I$ is the input image and $I_c$ is a color channel of $I$. $\boldsymbol{x}$ is one pixel in the image $I$ and $\Omega(\boldsymbol{x})$ is a local patch centered at $\boldsymbol{x}$. $\boldsymbol{y}$ is one neighboring pixel of $\boldsymbol{x}$. Note that the intensity values of all the images are normalized to 0-1 before calculating the loss.

We analyze the effectiveness of these three modifications in Section~\ref{enlighten}.

\subsection{Realistic Low-Light Mobile Photography Dataset}\label{dataset}
Current low-light enhancement benchmarks are mostly conducted on low-/high- light pairs captured by professional cameras (\eg, LOL~\cite{wei2018deep}). We observe several limitations on these benchmarks including 1) The collected data is mostly from a specific camera sensor, while the out-of-distribution generalization ability of a given approach is unable to be fairly evaluated.
2) There exists a huge gap between images captured by advanced professional cameras and smart phones.  The low-light images captured by phones typically suffer from much more ISO noise compared with images captured by professional cameras. Considering the above limitations, we propose to collect a new benchmark, named \textbf{R}ealistic \textbf{L}ow-light \textbf{M}obile \textbf{P}hotography (\textbf{RLMP}) dataset, to make up the shortcomings of current testbed.

Our main pipeline is described as following, we firstly fix the phone with a tripod and remotely control the camera by a program, in order to avoid physical interference during the process of capturing photos. The brightness of images is adjusted by changing exposure time and ISO. Similar to \cite{anaya2018renoir}, we adopt a pixel-alignment approach that helps eliminate the misalignment between the image pairs caused by camera shaking, object movement, lightness changing, and so on. Images are captured with three mobile phone models: Huawei Mate 10, Huawei P30, and Vivo NEX, and in a variety of scenes. Generally, the exposure of these captured images is uneven, leading to the image noise with non-uniform distributions that are more complex to tackle. In the end, we select 30 image pairs from over 100 captured image pairs. For each image, we convert it to PNG format with both high-resolution ($3952\times2960$) version and low-resolution ($800\times600$) version.

\begin{figure*}
\vspace{0.5cm}
\centering
\includegraphics[width=1.0\linewidth]{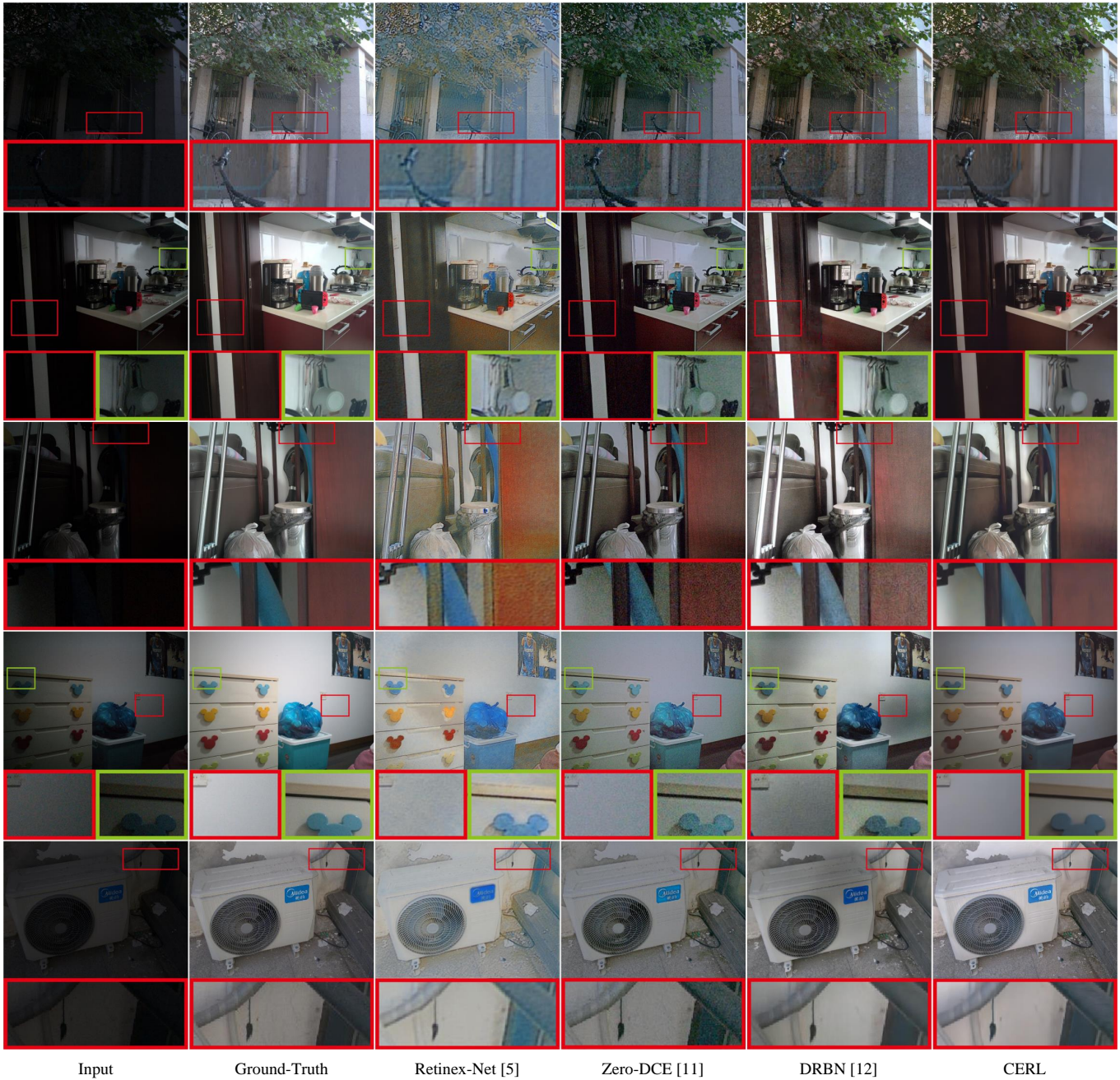}
\caption{Visual comparisons between the proposed method (CERL) and current state-of-the-art methods.}
\label{en1}
\end{figure*}

\section{Experiments}
\subsection{Experimental Settings}
\noindent{\textbf{Implementation.}}
We adopt the proposed EnlightenGAN+ as the light enhancement backbone and follow the specific training of EnlighteGAN~\cite{jiang2021enlightengan} while the denoiser backbone and pre-trained weight directly follow \cite{zhou2020awgn}.

In the self-supervised training of the denoiser, we start from the pre-trained model provided by the authors of \cite{zhou2020awgn} and then finetune the network for 5 epochs with a learning rate of 2e-6. An Adam optimizer with $\beta_1=0.9$ and $\beta_2=0.999$ and a batch size of 32 is adopted. For the CERL optimization framework, the total iteration number is set to 10, and $\mu$ is set to be increased from 0.1 to 0.9 uniformly.

\begin{table*}[t]
\small
{}
\caption{Quantitative evaluation on the LOL dataset.}
{\centering
\begin{tabular}{l|ccccccc}
\toprule
 Metric & LIME \cite{guo2016lime} & Retinex-Net \cite{wei2018deep}  & RRM \cite{li2018structure} & Zero-DCE \cite{guo2020zero} & EnlightenGAN \cite{jiang2021enlightengan} & DRBN \cite{yang2020fidelity} & CERL \\
\midrule
PSNR & 15.14 & 17.35  & 17.34 & 18.06 & 18.63 & {\color{blue}{20.17}} & {\color{red}{20.21}} \\
SSIM & 0.5051 & 0.7129 & 0.7226 & 0.5795 & 0.6767 & {\color{blue}{0.8122}} & {\color{red}{0.8154}}\\
NIQE & 9.6690 & {\color{blue}{3.9347}} & 4.8579 & 8.7667 & 5.5029 & 4.2476 & {\color{red}{3.8470}}\\
\bottomrule
\end{tabular}
\par}
\label{tablol}
\end{table*}

\begin{table*}[t]
\small
{}
\caption{Quantitative evaluation on our RLMP dataset.}
{\centering
\begin{tabular}{l|ccccccc}
\toprule
 Metric & LIME \cite{guo2016lime} & Retinex-Net \cite{wei2018deep} & RRM \cite{li2018structure} & Zero-DCE \cite{guo2020zero} & EnlightenGAN \cite{jiang2021enlightengan} & DRBN \cite{yang2020fidelity} & CERL \\
\midrule
PSNR & 11.11 & 18.26 & 18.25 & 17.86 & 18.22 & {\color{blue}{18.68}} & {\color{red}{20.05}} \\
SSIM & 0.5605 & 0.6984 & 0.7033 & 0.5892 & 0.5326 & {\color{blue}{0.8094}} & {\color{red}{0.8129}} \\
NIQE & 5.4099 & 4.1748 & 3.2010 & 3.5459 & 4.4779 & {\color{red}{2.7830}} & {\color{blue}{3.1352}} \\
\bottomrule
\end{tabular}
\par}
\label{tabrlmp}
\end{table*}

\noindent{\textbf{Dataset.}}
We take the unpaired training set from EnlightenGAN~\cite{jiang2021enlightengan} to train our light enhancement backbone. For the self-supervised fine-tuning of denoiser, we adopt low-light noisy images from the training set of LOL dataset~\cite{wei2018deep}. We take paired low-/high- light images from the test set of LOL and RLMP datasets for testing. For LOL dataset, we follow the dataset partition scheme in DRBN \cite{yang2020fidelity}, where the training set includes 689 low-/high- light image pairs and testing set includes 100 image pairs. For RLMP dataset, we use images of the low-resolution version for all experiments. 

\subsection{Results of CERL}
We compare CERL with state-of-the-art low-light enhancement methods, including LIME \cite{guo2016lime}, Retinex-Net \cite{wei2018deep}, Refined Retinex Model (RRM) \cite{li2018structure}, Zero-DCE \cite{guo2020zero}, EnlightenGAN \cite{jiang2021enlightengan} and DRBN \cite{yang2020fidelity}. For Retinex-Net and DRBN, we train the models from stretch on the training set of LOL dataset mentioned above that includes 689 low-/high- light image pairs. For Zero-DCE and EnlightenGAN, we train them on the same data settings proposed in their original papers. We make sure that the images for testing are unseen to all the training processes of the selected models for comparison. We adopt three objective evaluation metrics: Peak Signal-to-Noise Ratio (PSNR), Structural SIMilarity (SSIM) \cite{wang2004image}, and Natural Image Quality Evaluator (NIQE) \cite{mittal2012making}. The first two metrics are full-referenced image quality assessments while the third one is a no-referenced assessment which evaluates human visual perception quality of images without reference. 

\textbf{Quantitative Evaluations.}
As shown in Table~\ref{tablol} and \ref{tabrlmp}. CERL reaches the best scores in terms of both PSNR and SSIM metrics and also performs well on NIQE, demonstrating the superiority of the proposed method. In addition, DRBN \cite{yang2020fidelity} shows comparably good performance on LOL test set, while its performance degrades on RLMP dataset, indicating that models trained on a specific noise domain may fail to generalize to another domain. Others show unsatisfactory results due to the poor performance on noise removal (Zero-DCE, LIME, and EnlightenGAN) or generate texture artifacts during the light enhancement (Retinex-Net). In contrast, CERL shows impressive results on both two datasets thanks to the self-supervised training scheme and modified light enhancement backbone.

\textbf{Qualitative Evaluations.}
Fig.~\ref{en1} shows the qualitative results on the testing sets. We can see that Retinex-Net \cite{wei2018deep} produces over-exposed and unrealistic results. Zero-DCE \cite{guo2020zero} well restores the global illumination, but it does not take noise into consideration, hence the low-light noise is amplified in the results. DRBN \cite{yang2020fidelity} performs well in suppressing the noise, but fails to handle those regions with heavy noise and also tends to produce artifacts. Compared with them, the results of CERL are of visually pleasing illumination as well as clean textural details without noise and artifacts.

\begin{figure}
\centering
\subfigure[Before finetuning]{
\begin{minipage}[b]{0.475\linewidth}
\includegraphics[width=1.03\linewidth]{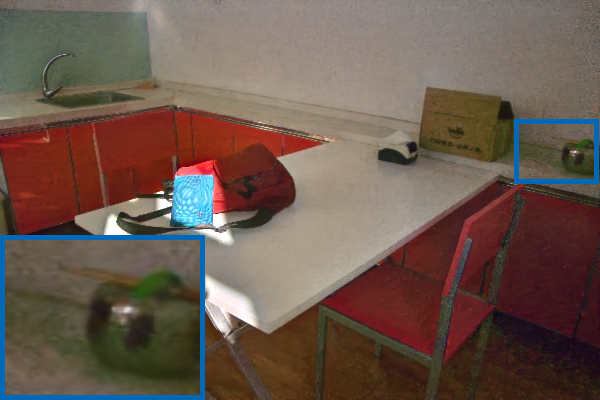}
\end{minipage}}
\subfigure[After finetuning]{
\begin{minipage}[b]{0.475\linewidth}
\includegraphics[width=1.03\linewidth]{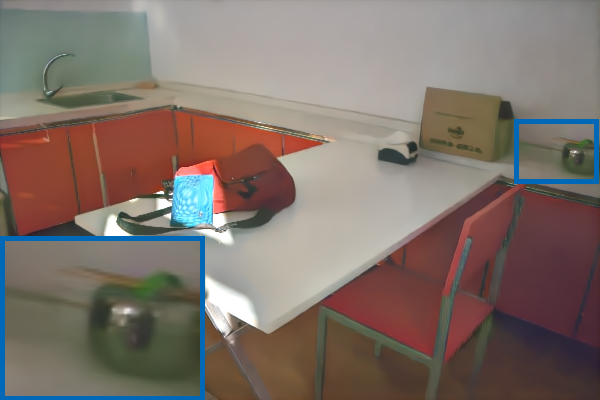}
\end{minipage}}
\caption{Visual comparison for denoiser fine-tuning.}
\label{gen1}
\end{figure}

\begin{table}
\small
{}
\caption{Quantitative evaluation for denoiser fine-tuning.}
\vspace{-0.1cm}
{\centering
\begin{tabular}{l|cc|cc}
\toprule
Dataset& \multicolumn{2}{c|}{LOL}  & \multicolumn{2}{c}{RLMP} \\ 
Metric &  PSNR & SSIM & PSNR & SSIM\\
\toprule
Noisy input & 19.58 & 0.6969 & 18.22 & 0.5613\\
w/ Pre-trained & {\color{blue}{19.66}} & {\color{blue}{0.7705}} & {\color{blue}{18.48}} & {\color{blue}{0.6754}}\\
w/ Fine-tuned & {\color{red}{20.07}} & {\color{red}{0.8064}} & {\color{red}{20.00}} & {\color{red}{0.7602}} \\ 
\bottomrule
\end{tabular}
\par}
\label{tab1}
\end{table}


\begin{figure*}
\centering
\subfigure[Enhanced]{
\begin{minipage}[b]{0.185\linewidth}
\includegraphics[width=1.065\linewidth]{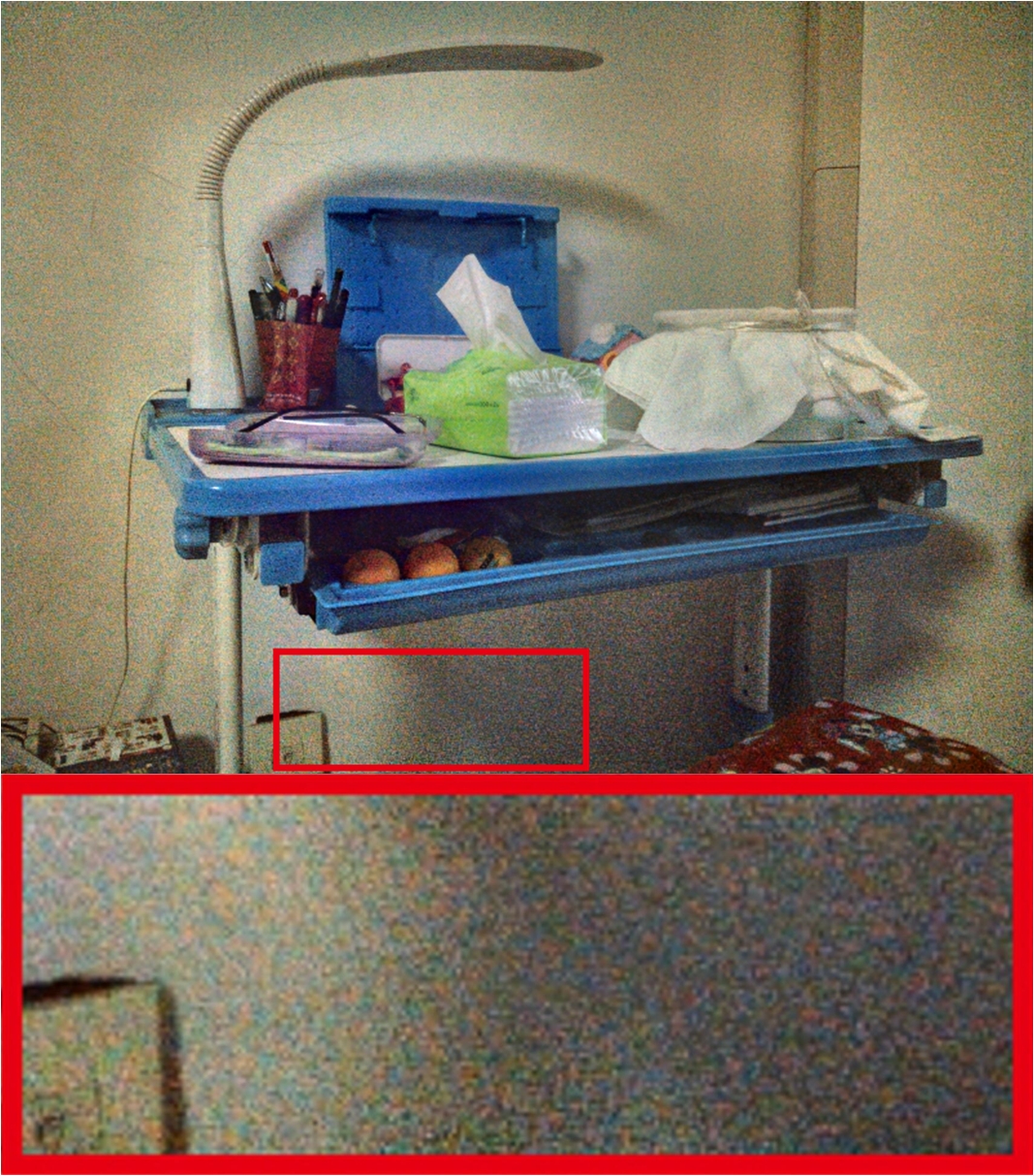}
\includegraphics[width=1.065\linewidth]{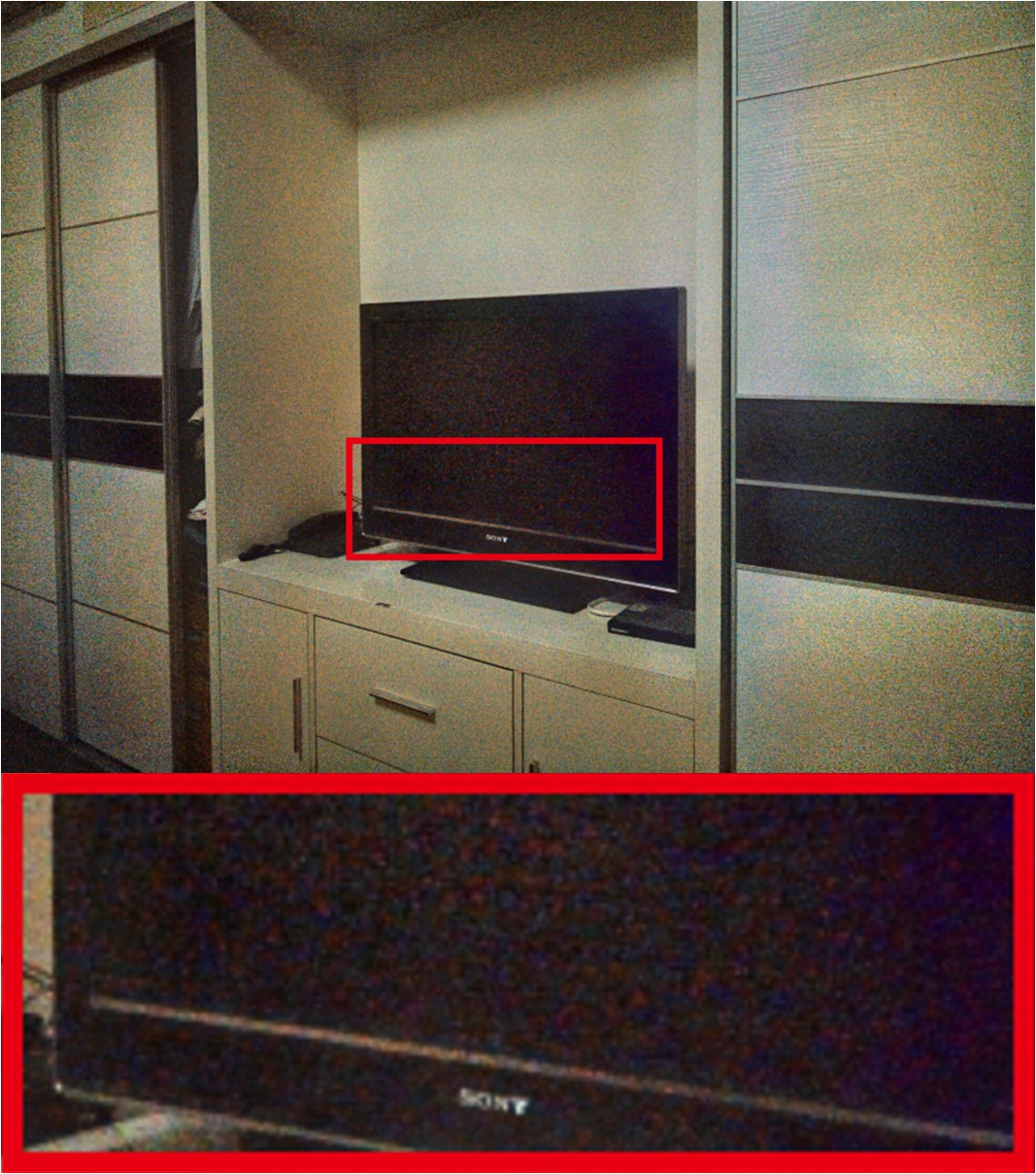}
\end{minipage}}
\subfigure[+BM3D \cite{dabov2007image}]{
\begin{minipage}[b]{0.185\linewidth}
\includegraphics[width=1.065\linewidth]{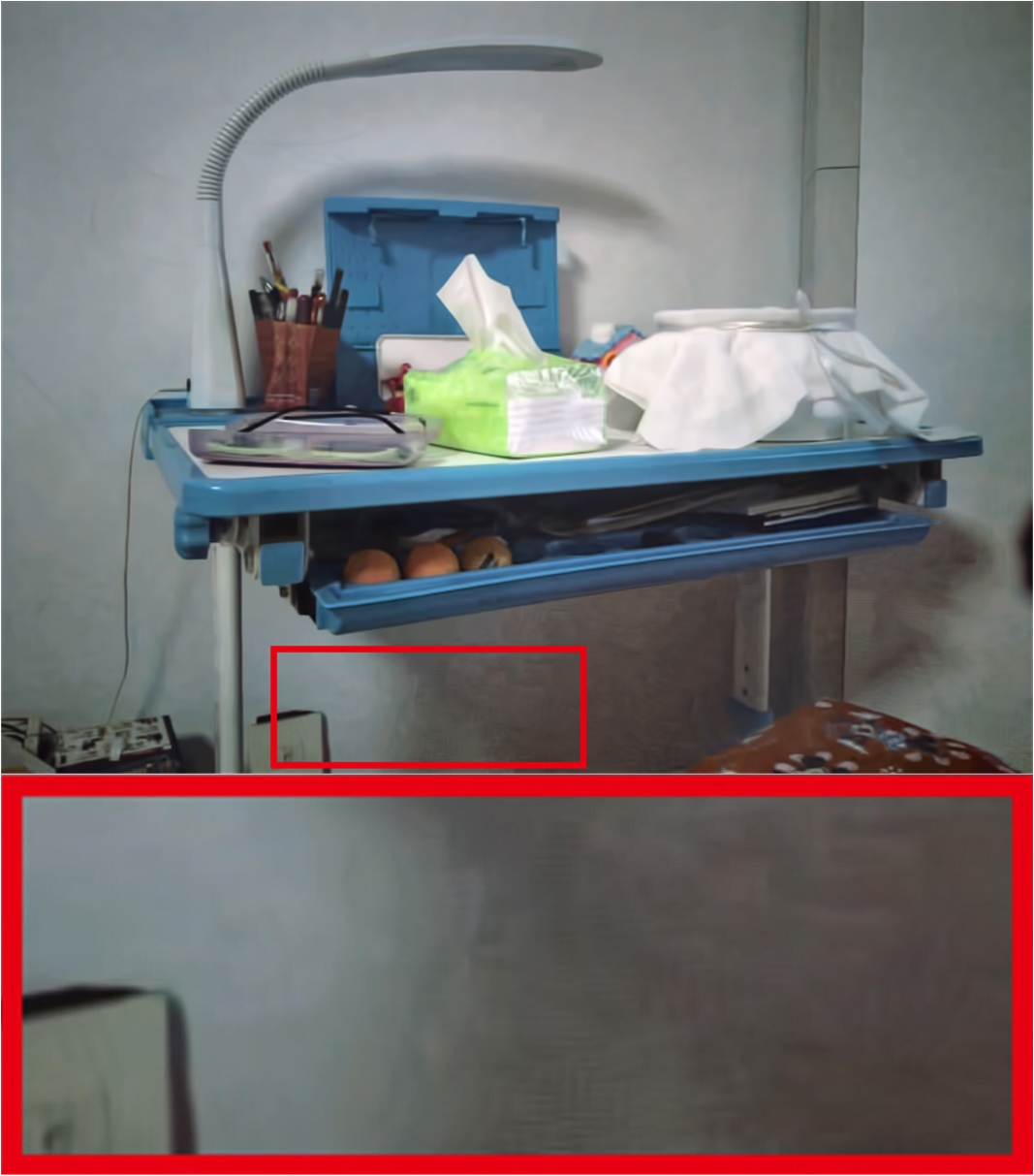}
\includegraphics[width=1.065\linewidth]{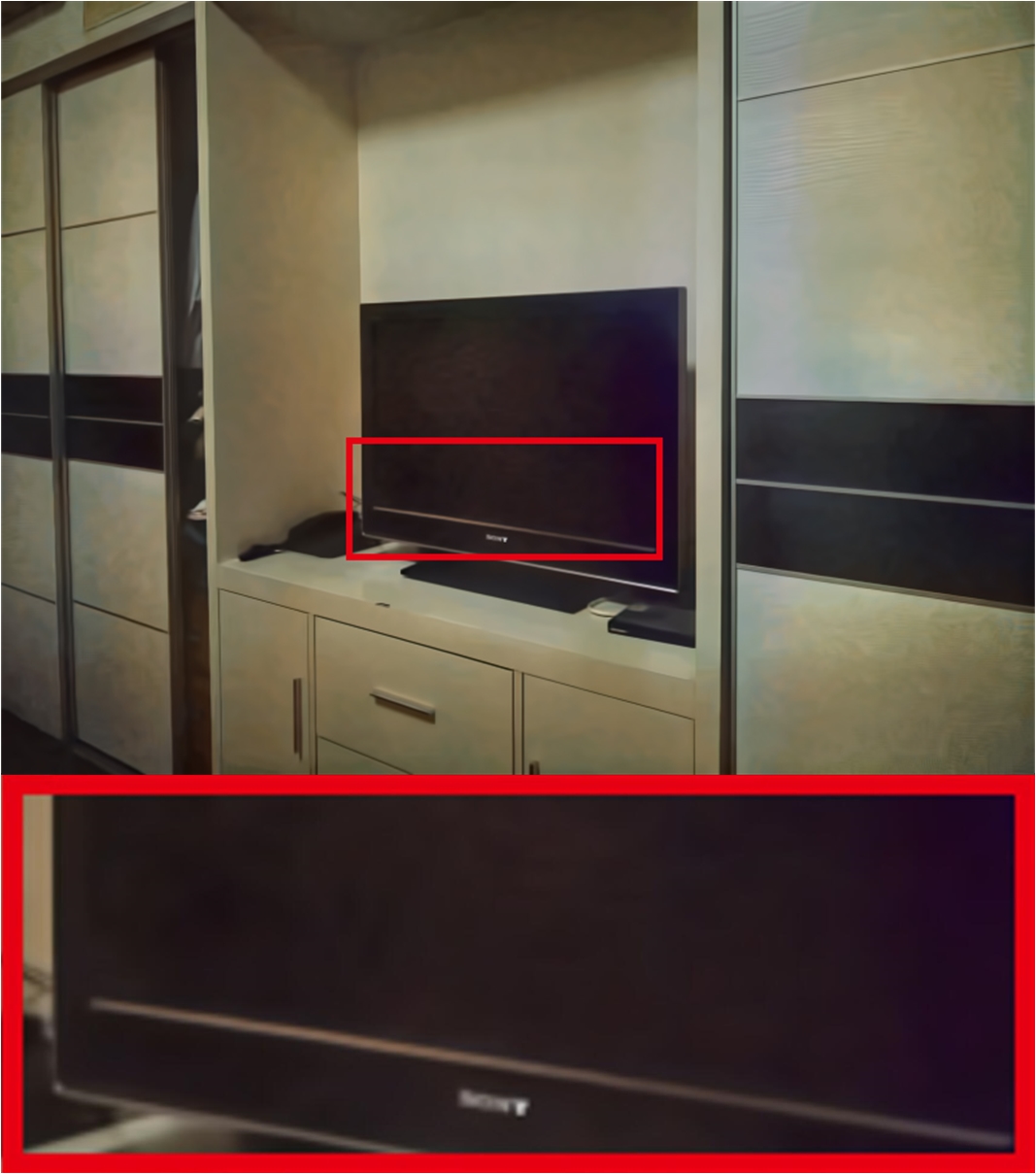}
\end{minipage}}
\subfigure[+DBSN \cite{wu2020unpaired}]{
\begin{minipage}[b]{0.185\linewidth}
\includegraphics[width=1.065\linewidth]{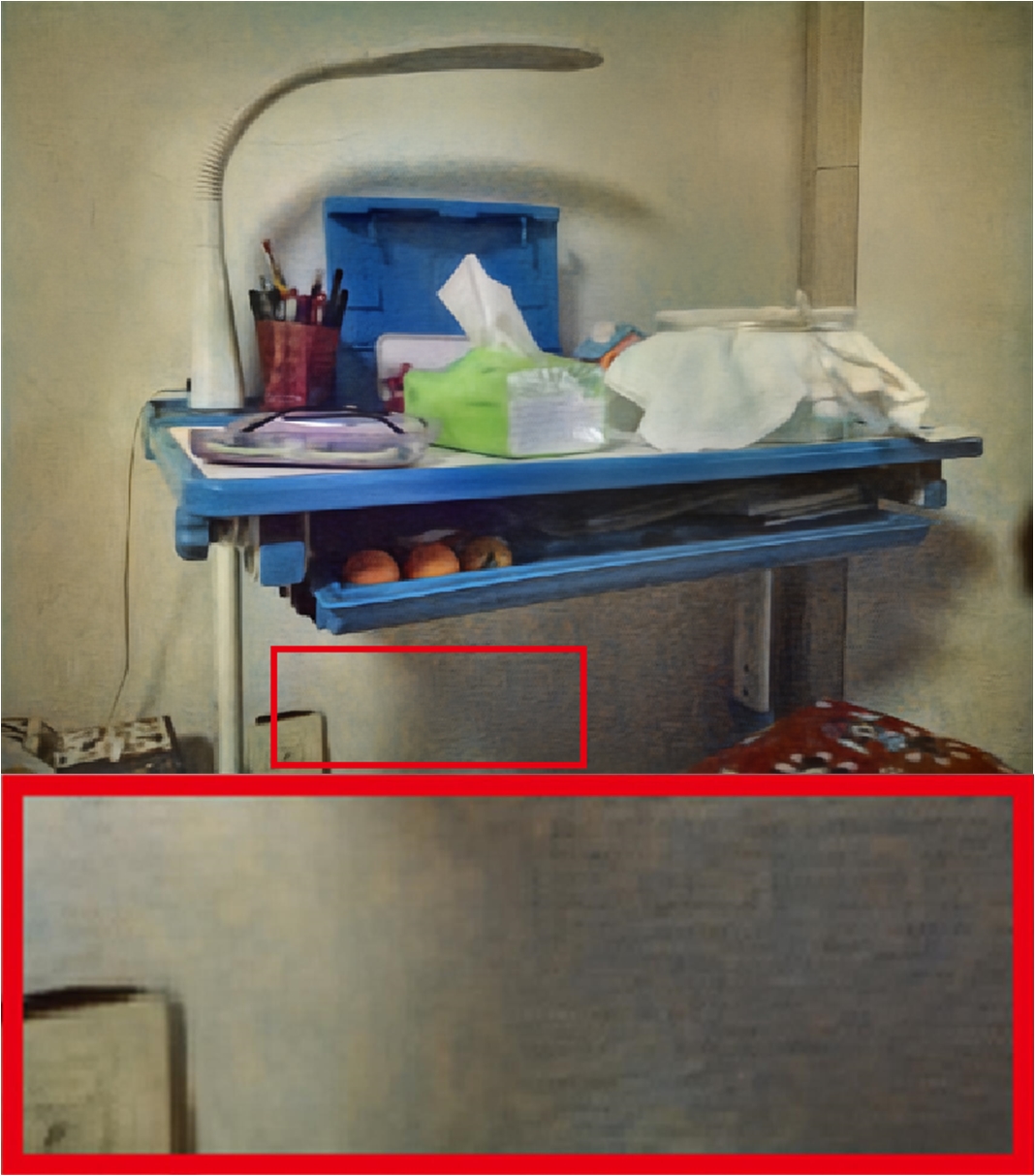}
\includegraphics[width=1.065\linewidth]{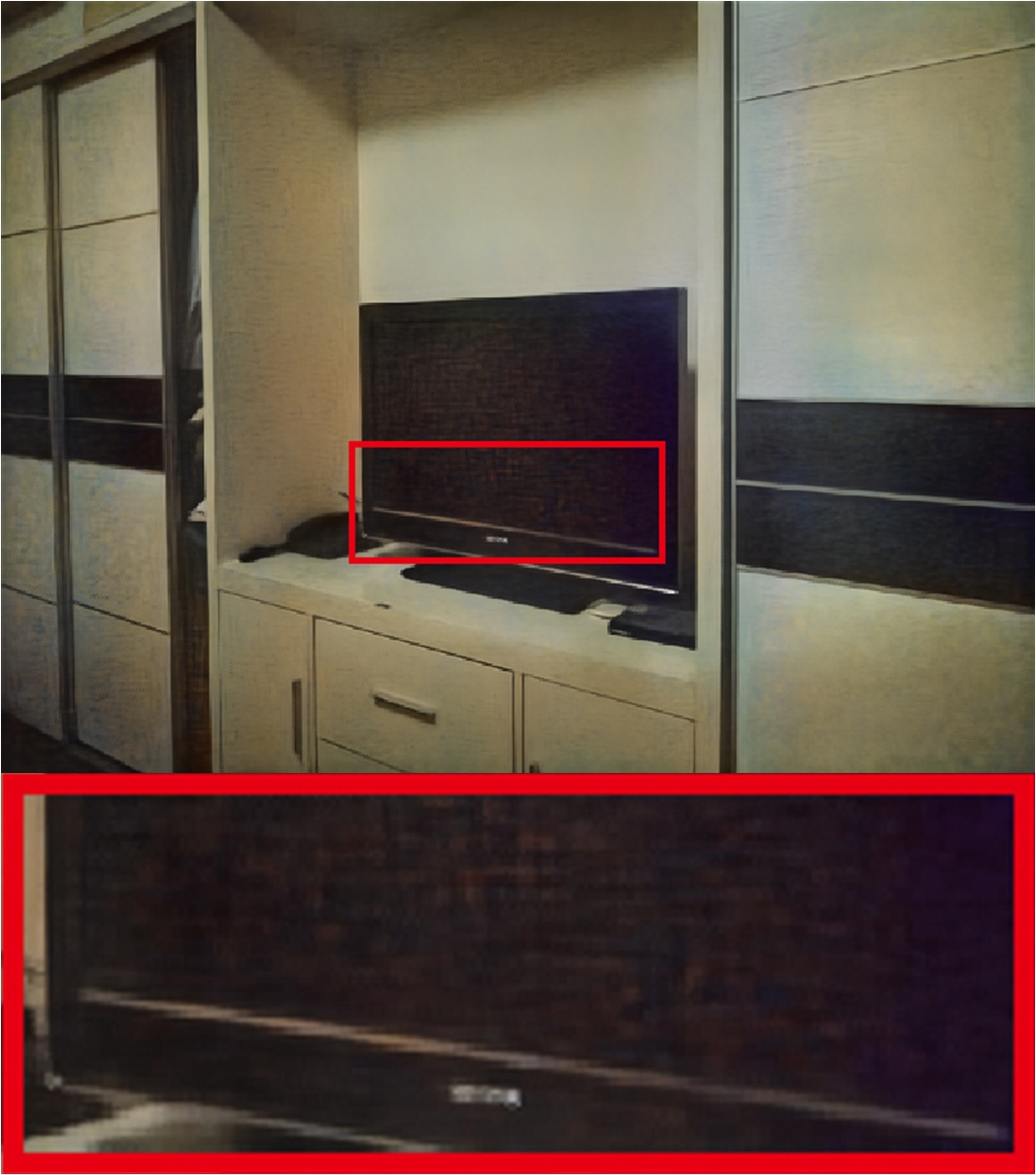}
\end{minipage}}
\subfigure[+Fine-tuned]{
\begin{minipage}[b]{0.185\linewidth}
\includegraphics[width=1.065\linewidth]{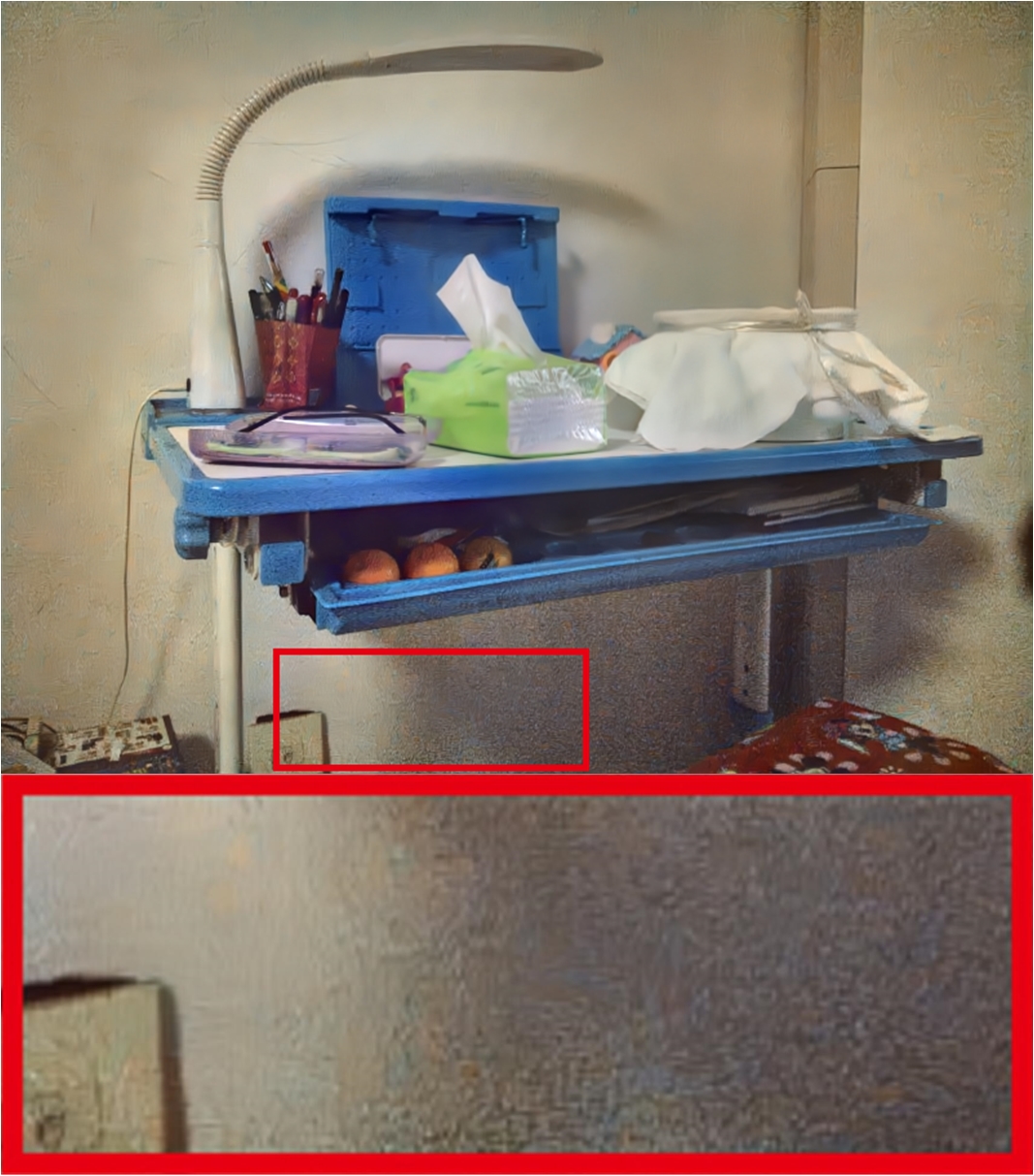}
\includegraphics[width=1.065\linewidth]{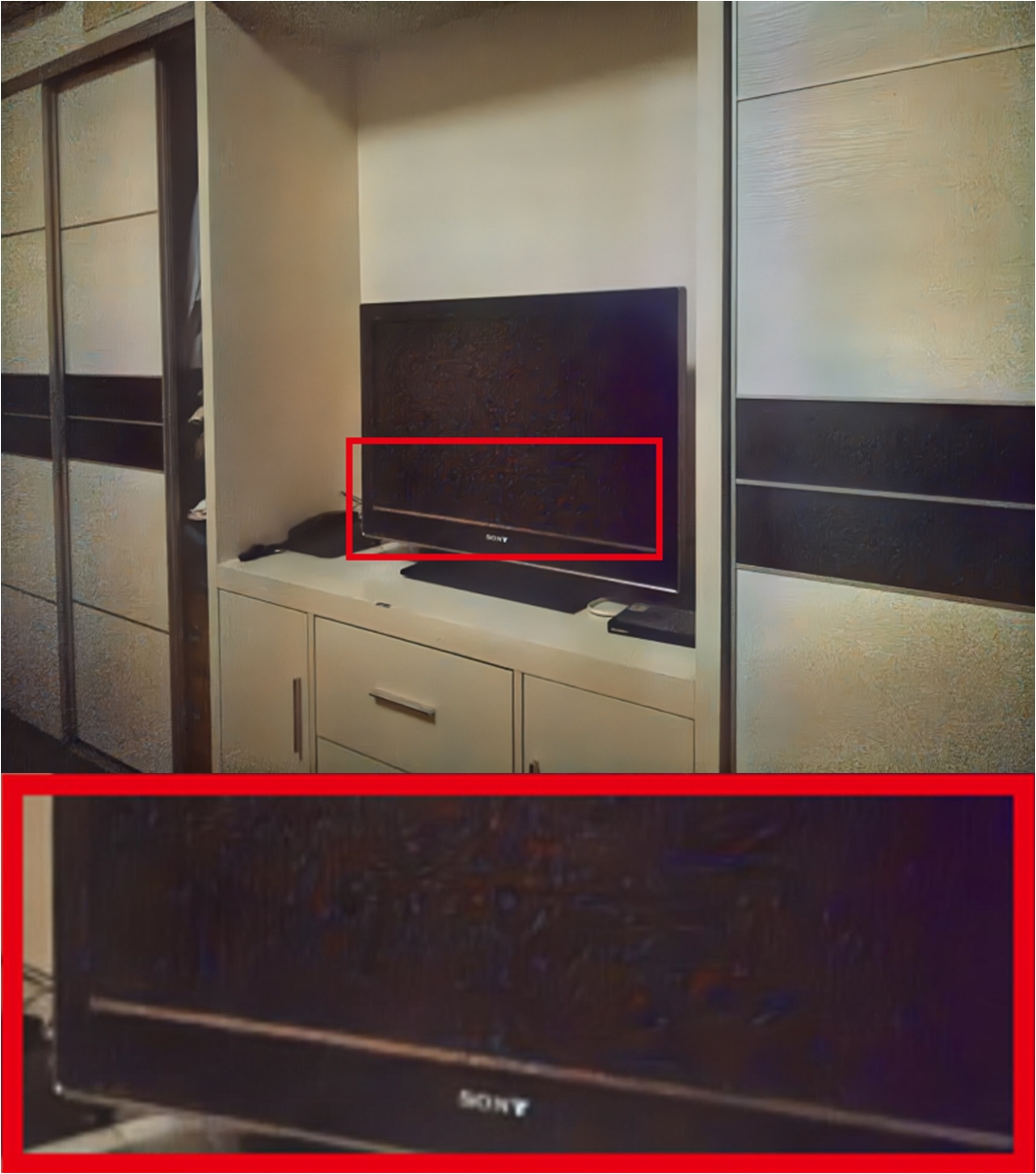}
\end{minipage}}
\subfigure[CERL]{
\begin{minipage}[b]{0.185\linewidth}
\includegraphics[width=1.065\linewidth]{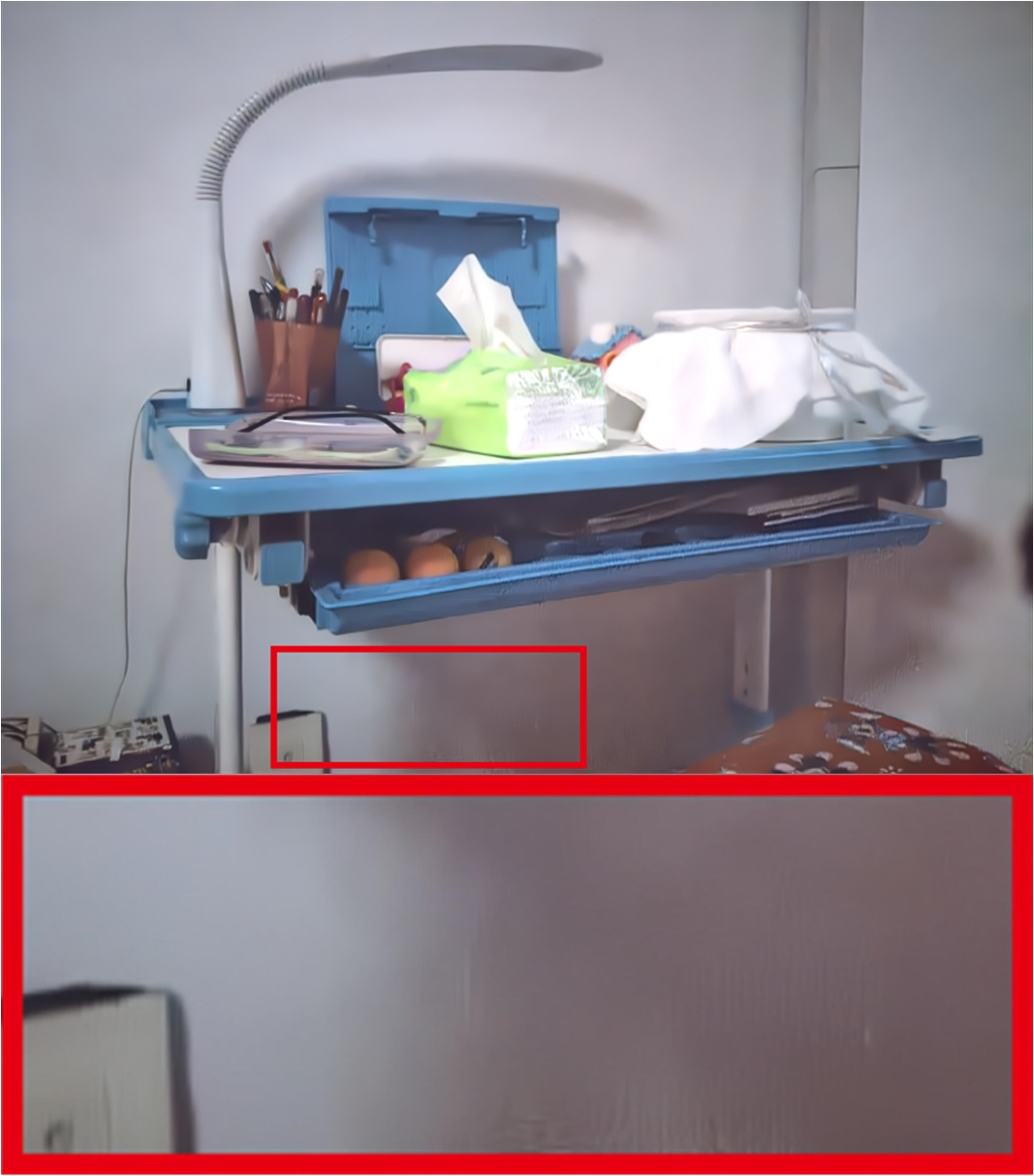}
\includegraphics[width=1.065\linewidth]{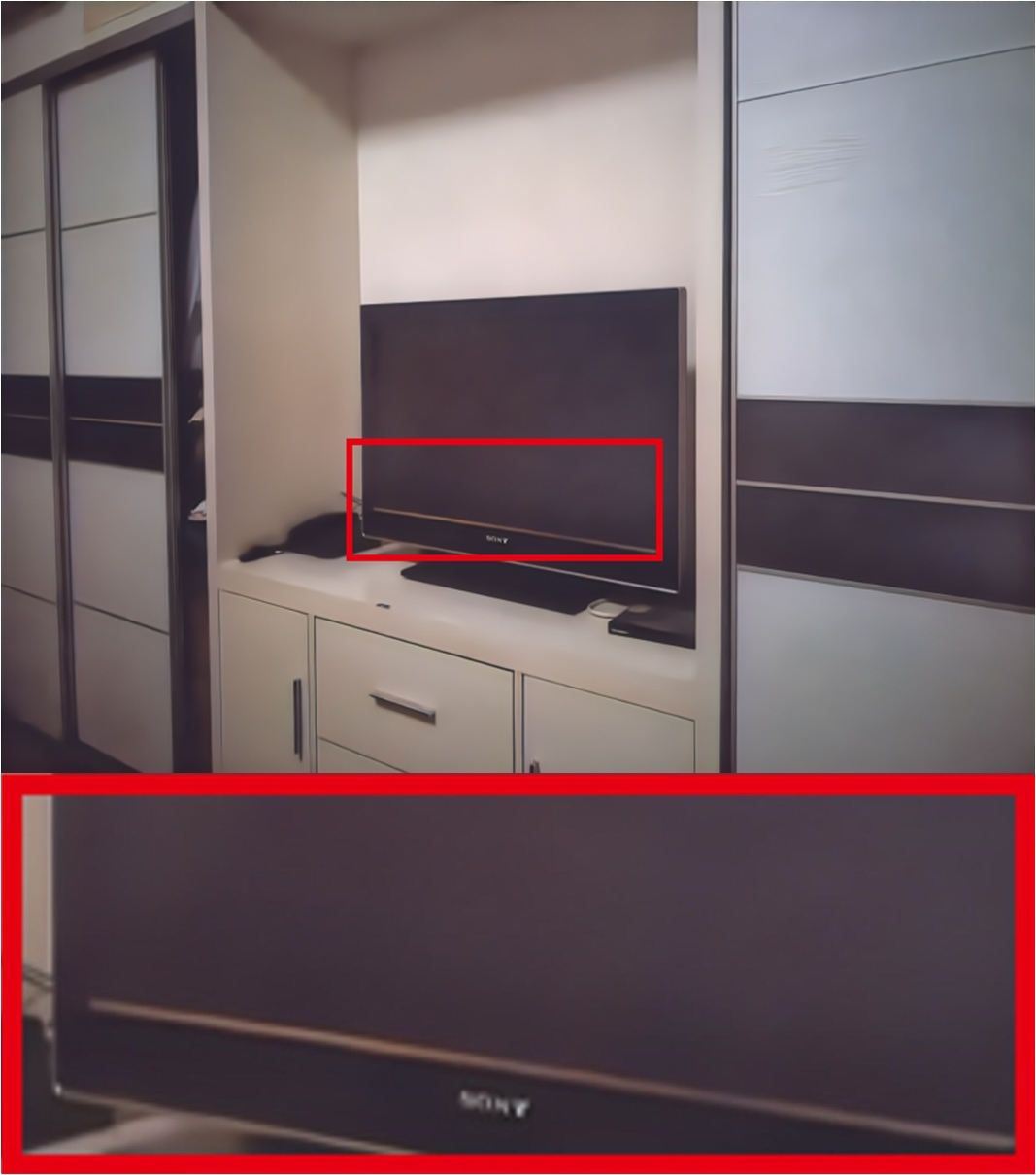}

\end{minipage}}

\caption{Visual comparison of the denoising performance for real-world low-light noise.}
\label{pnp1}
\end{figure*}

\begin{figure*}
\centering
\subfigure[Iteration 1]{
\begin{minipage}[b]{0.237\linewidth}
\includegraphics[width=1.05\linewidth]{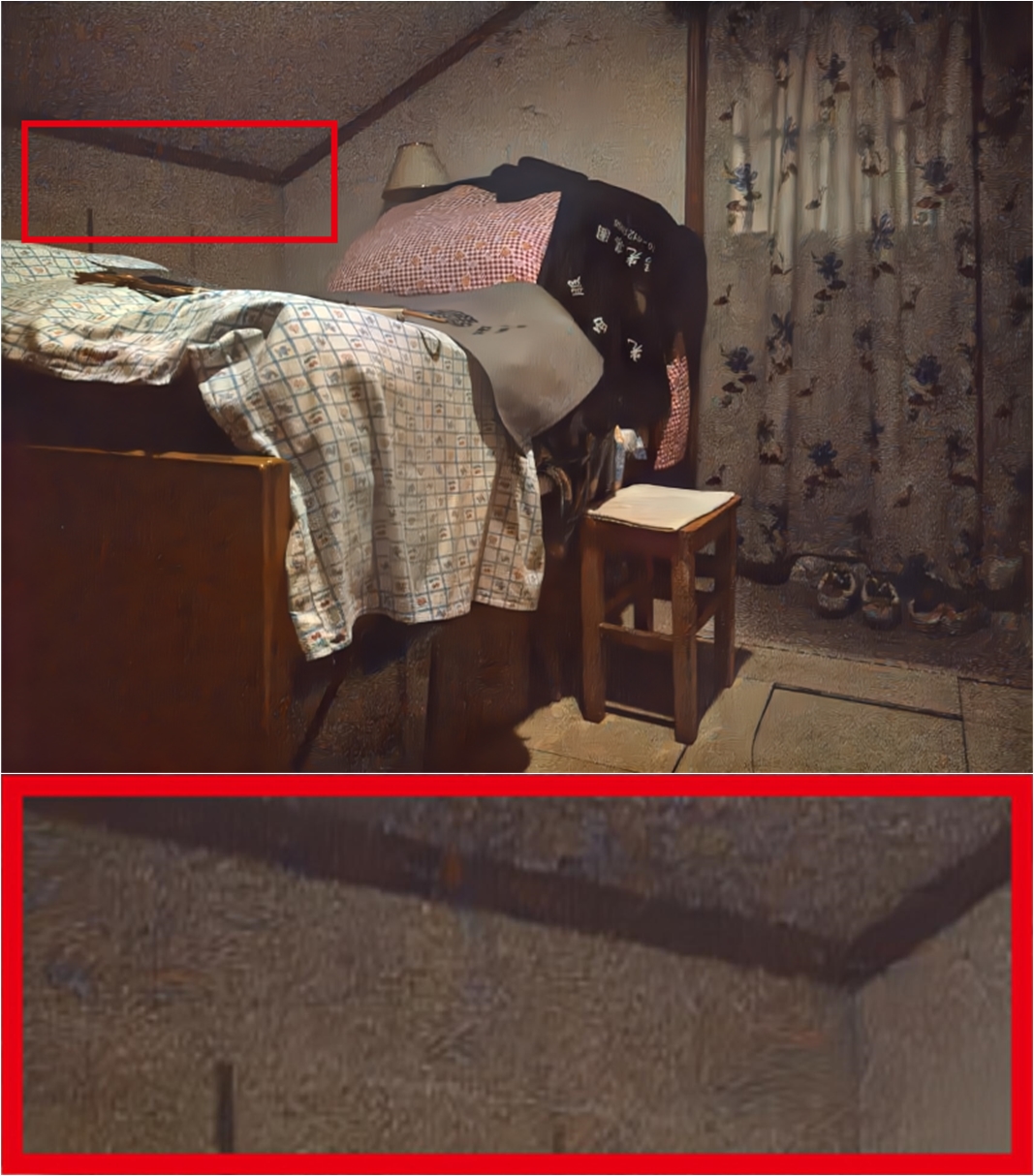}
\end{minipage}}
\subfigure[Iteration 4]{
\begin{minipage}[b]{0.237\linewidth}
\includegraphics[width=1.05\linewidth]{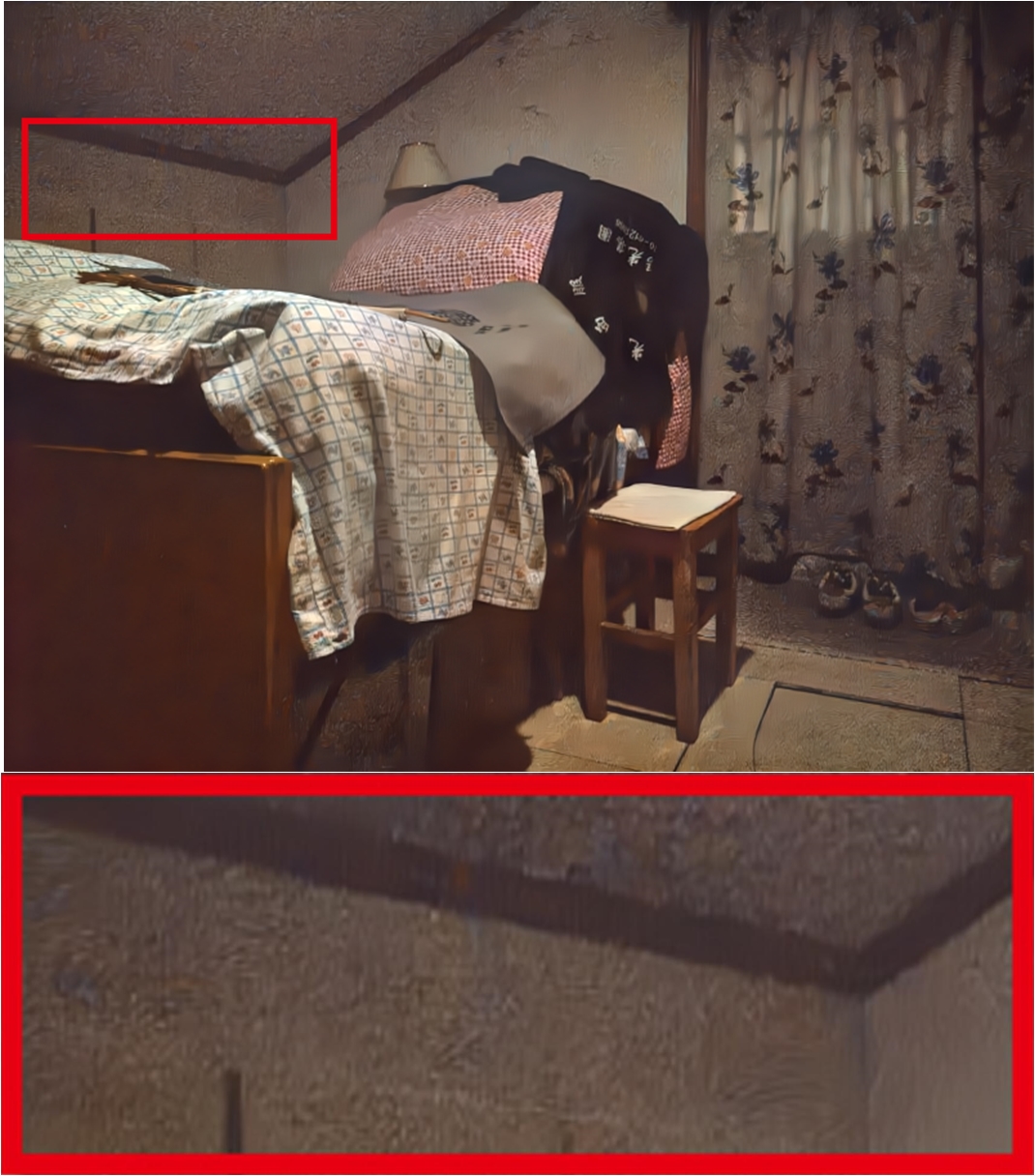}
\end{minipage}}
\subfigure[Iteration 7]{
\begin{minipage}[b]{0.237\linewidth}
\includegraphics[width=1.05\linewidth]{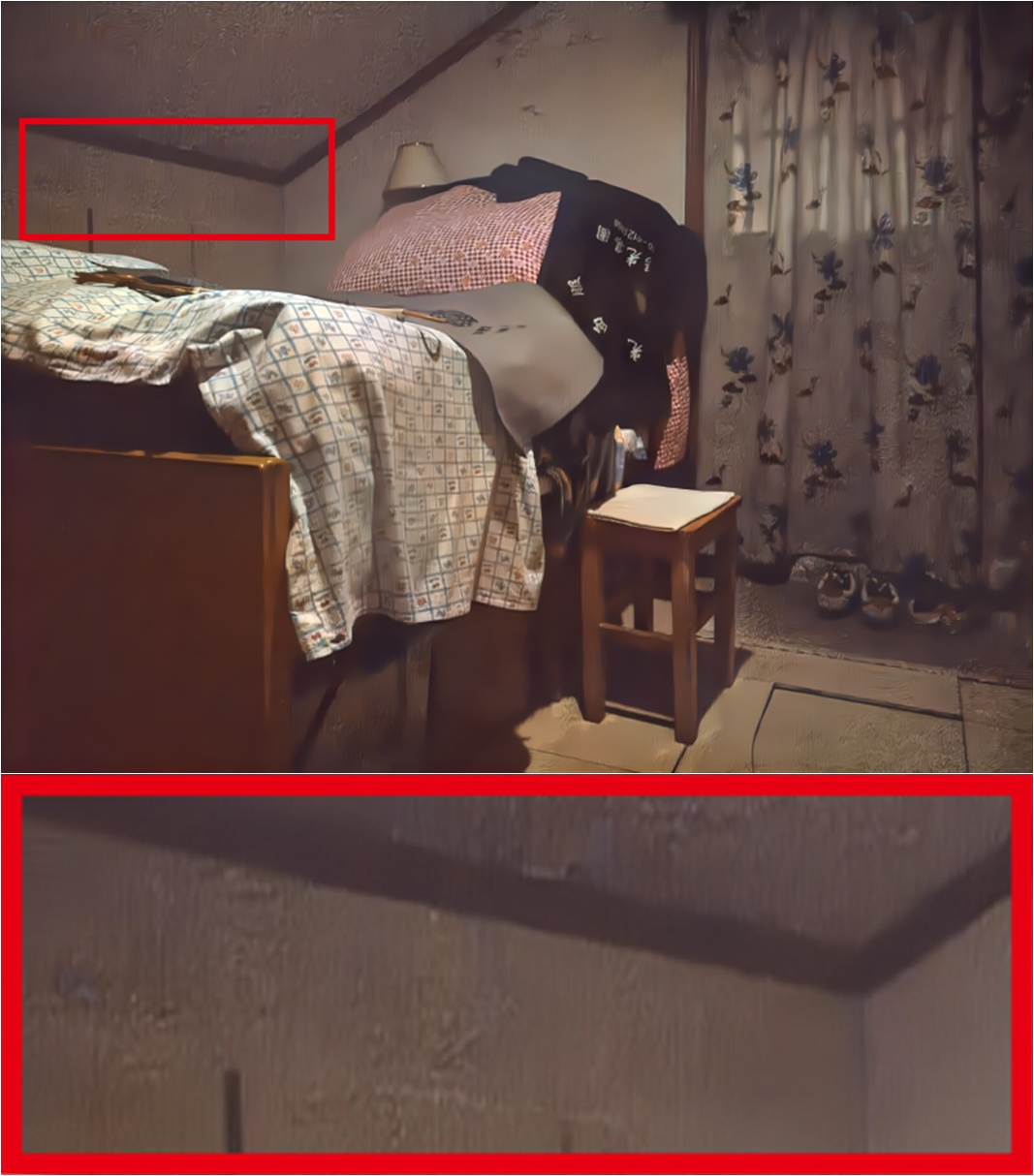}
\end{minipage}}
\subfigure[Iteration 10]{
\begin{minipage}[b]{0.237\linewidth}
\includegraphics[width=1.05\linewidth]{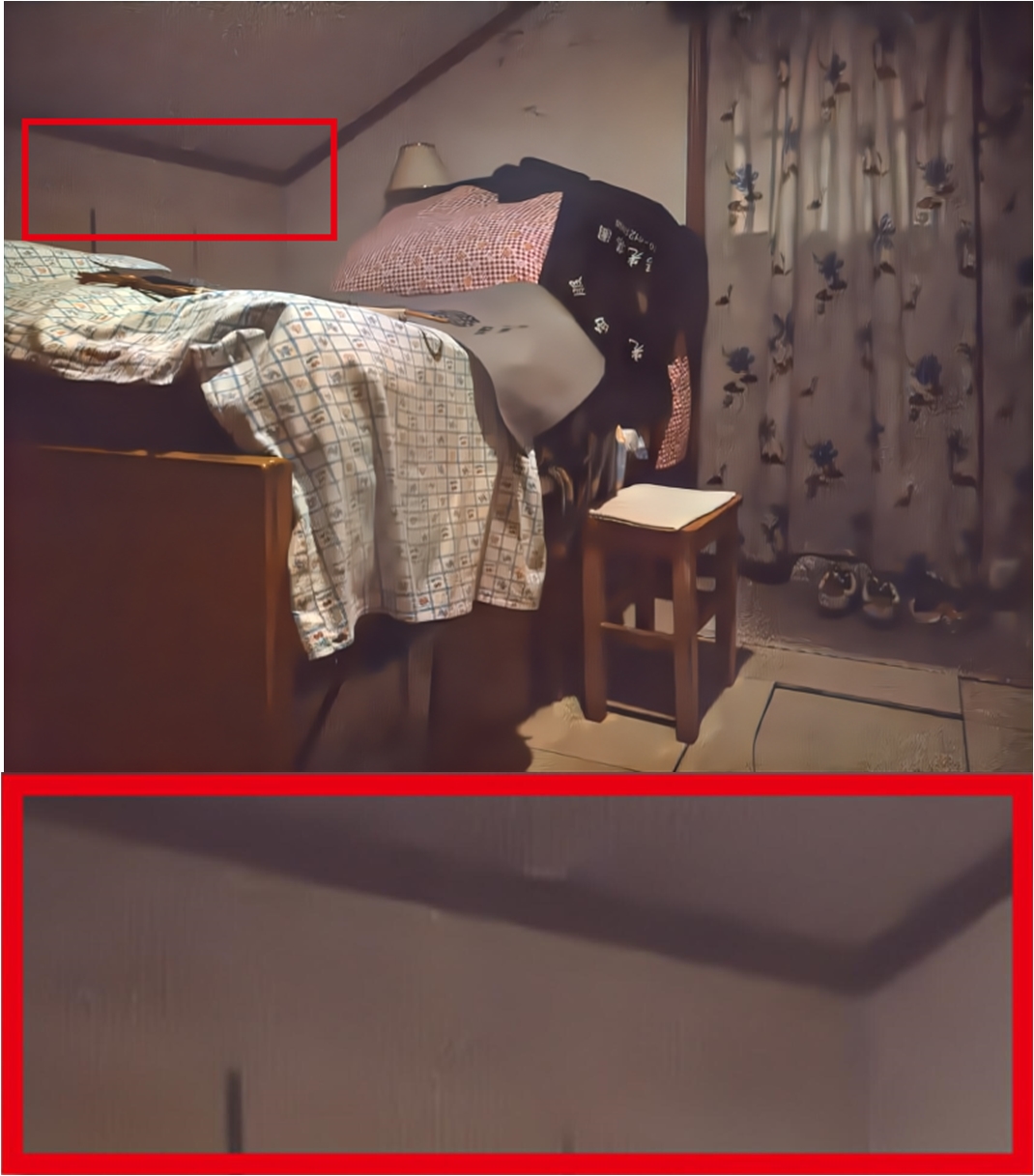}
\end{minipage}}
\caption{Intermediate results of $\boldsymbol{x}$ at different iterations of CERL.}
\label{pnp2}
\end{figure*}

\begin{figure}
\centering
\subfigure{
\begin{minipage}[b]{0.475\linewidth}
\includegraphics[width=1.03\linewidth]{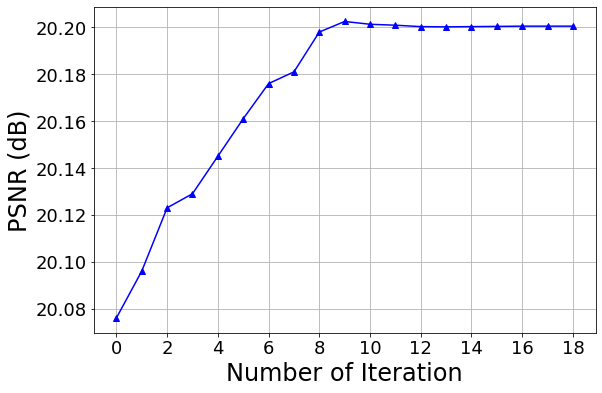}
\end{minipage}}
\subfigure{
\begin{minipage}[b]{0.475\linewidth}
\includegraphics[width=1.03\linewidth]{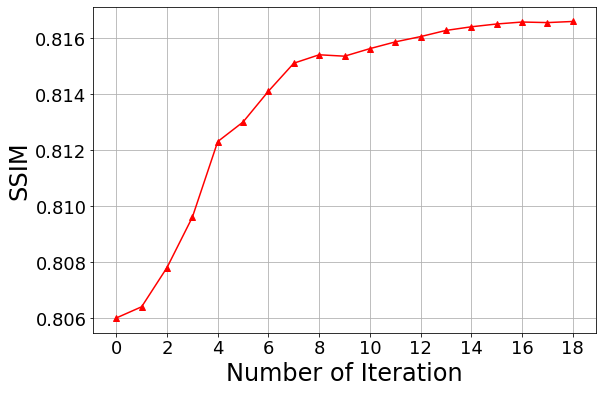}
\end{minipage}}
\caption{PSNR and SSIM scores on LOL dataset during iterations.}
\label{gru}
\end{figure}

\begin{table}
\small
{}
\caption{Quantitative evaluation for denoising performance.}
{\centering 
\begin{tabular}{l|cc|cc}
\toprule
Dataset& \multicolumn{2}{c|}{LOL}  & \multicolumn{2}{c}{RLMP} \\
Metric &  PSNR & SSIM & PSNR & SSIM\\
\toprule
Noisy input & 19.58 & 0.6969 & 18.22 & 0.5613\\
w/ BM3D \cite{dabov2007image} & 19.76 & 0.7939 & 18.60 & 0.7578 \\
w/ DBSN \cite{wu2020unpaired} & 19.71& 0.7758 & 18.46 & 0.7302 \\
w/ Fine-tuned & {\color{blue}{20.07}} & {\color{blue}{0.8064}} & {\color{blue}{20.00}} & {\color{blue}{0.7602}} \\ 
CERL & {\color{red}{20.21}} & {\color{red}{0.8154}} & {\color{red}{20.05}} & {\color{red}{0.8129}} \\
\bottomrule
\end{tabular}
\par}
\label{tabde}
\end{table}

\subsection{Performance of Fine-Tuned Denoiser.}
To demonstrate the superiority of our self-supervised training scheme, we compare the performance of the denoiser before and after fine-tuning. Images enhanced by our light enhancement backbone are treated as noisy input. The quantitative results are shown in Table \ref{tab1}. One can see that the fine-tuned denoiser achieves better performance in PSNR and SSIM on both datasets. Fig.~\ref{gen1} presents visual comparisons. While the result of the pre-trained denoiser is still noisy (see the zoom-in part), the fine-tuned denoiser succeeds in removing the remaining noise as well as preserving sharp edges and textural details.

\subsection{More visual comparisons of CERL.}
\textbf{Performance boosting on denoising.}
We illustrate that the proposed framework boosts the denoising performance in the low-light enhancement task. Compared with the widely used pipeline that adopts an off-the-shelf denoiser for post-processing, our framework makes two main changes: 1) fine-tuning the denoiser on low-light noise, 2) the CERL optimization framework. We conduct experiments to demonstrate that both changes make solid contributions for improving denoising performance. We use images enhanced by our enhancement backbone as the noisy input. For comparison, we take two state-of-the-art denoisers, BM3D \cite{dabov2007image} and DBSN \cite{wu2020unpaired}, together with our fine-tuned denoising model, all as the post-processing denoising methods. Quantitative results are shown in Table \ref{tabde}. One can see that as the fine-tuned denoising model achieves better PSNR and SSIM compared with state-of-the-art methods, the proposed CERL further boosts the performance on both PSNR and SSIM. Fig.~\ref{pnp1} provides the visual results. We can observe that the results of BM3D are over-smoothed and lose details. DBSN preserves details and sharp edges, but suffers from artifacts in those regions having heavy noise. Our fine-tuned model does well in removing noise but fails to handle extremely noisy regions, still generating subtle artifacts in its results. CERL outperforms all the other methods, of which the results have smooth background and fine textural details, without undesirable artifacts. In addition, the original enhancement results are clearly suffering from color distortion, and CERL helps restore the correct colors and further improve the light enhancement quality.

\begin{table*}[h]
    \centering
    \caption{Ablation study for modules in EnlightenGAN+ on LOL \cite{wei2018deep}.}
    \setlength{\tabcolsep}{1.3mm}
    \begin{tabular}{c|ccccccc}
    \toprule
    Modules & & & & \\
    \midrule
    Bright Channel Loss & & \checkmark & &  & \checkmark & \checkmark & \checkmark \\
    Channel Attention & &  & \checkmark & & \checkmark&  & \checkmark\\
    Pyrimid Pooling & & & & \checkmark & &\checkmark & \checkmark\\
    \midrule
    PSNR & 18.63 & 19.09 & 18.65 & 18.83 & 19.21 & 19.16 & 19.58 \\
    SSIM & 0.6767 & 0.6845 & 0.6762 & 0.6907 & 0.6912 & 0.6817 & 0.6969\\
    \bottomrule
    \end{tabular}
    \label{tab_ablation_enlightengan}
\end{table*}

\textbf{Intermediate Results of CERL.}
Fig.~\ref{pnp2} presents the visual results of $\boldsymbol{x}$ at different iterations of CERL. As the iteration number increases, the noise is more removed. Besides, the illumination of the enhanced results is also refined during iterations.

\textbf{Performance Curve.}
We provide the performance curves of CERL on PSNR and SSIM  in Fig.~\ref{gru}. To demonstrate the tendency more clearly, we extend the iteration number to 20 and fix the penalty parameter to 0.9 after the ninth iteration. In the results, PSNR and SSIM scores both become stable and gradually converge after fixing the penalty number.


\subsection{Ablation Study for EnlightenGAN+}\label{enlighten}
We conduct several ablation studies to show the effectiveness of our modifications on EnlightenGAN \cite{jiang2021enlightengan}. The three modifications in the proposed EnlightenGAN+ are: 1) bright channel loss, 2) channel attention, and 3) pyramid pooling module. Quantitative results on LOL dataset are in Table \ref{tab_ablation_enlightengan}. The results show that the three modifications are compatible with each other, and their combination to further improvement. Note that single channel attention does not improve model performance, while the combination of bright channel loss and channel attention leads to performance improvement. We attribute this result to the different channel information properties between models. The information in feature channels of the original model does not contain any specific inductive bias, while for the model trained with bright channel loss, different feature channels typically contain different parts of information that are effective for light enhancement. Therefore, a channel attention module could explicitly guide the model to focus on the most informative channels.

\section{Conclusion}
In this paper, we introduce Coordinated Enhancement with Real Low-light noise (CERL), which disentangles the low-light enhancement task into two sub-problems, namely light enhancement and noise removal, and integrates them into a unified optimization framework. Under the framework, we present a self-supervised fine-tuning scheme to obtain a deep denoiser that is easily adapted for real low-light noise; as well as an improved light enhancement backbone. Experiments demonstrate the superiority of our framework against other state-of-the-art methods.  

{\small
\bibliographystyle{IEEEtran}
\bibliography{egbib}
}

\begin{IEEEbiography}[{\includegraphics[width=1in,height=1.25in,clip,keepaspectratio]{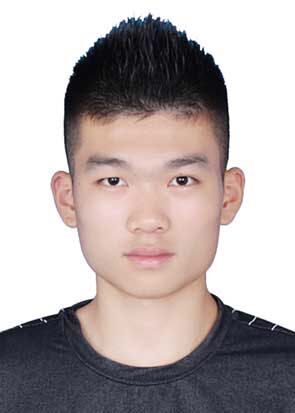}}]{Zeyuan Chen}
is currently an undergraduate student at University of Science and Technology of China. His research interests lie on deep learning and computer vision.
\end{IEEEbiography}

\begin{IEEEbiography}[{\includegraphics[width=1in,height=1.25in,clip,keepaspectratio]{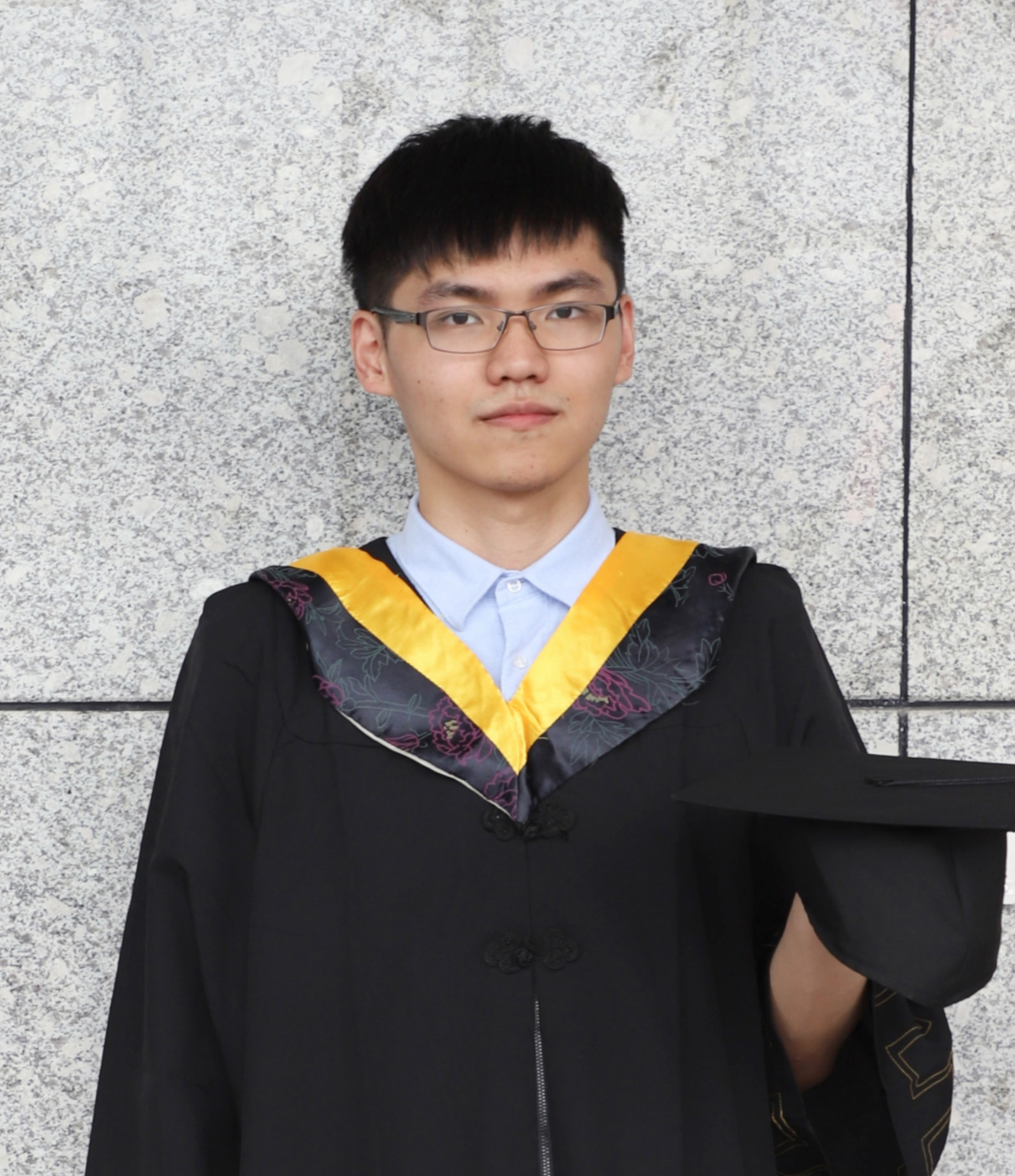}}]{Yifan Jiang}
is currently a Ph.D. student at The University of Texas at Austin. He received his bachelor degree from the Huazhong University of Science and Technology in 2019. His research interests are on computer vision and deep learning, in particular generative models, neural architecture search, and representation Learning.
\end{IEEEbiography}

\begin{IEEEbiography}[{\includegraphics[width=1in,height=1.25in,clip,keepaspectratio]{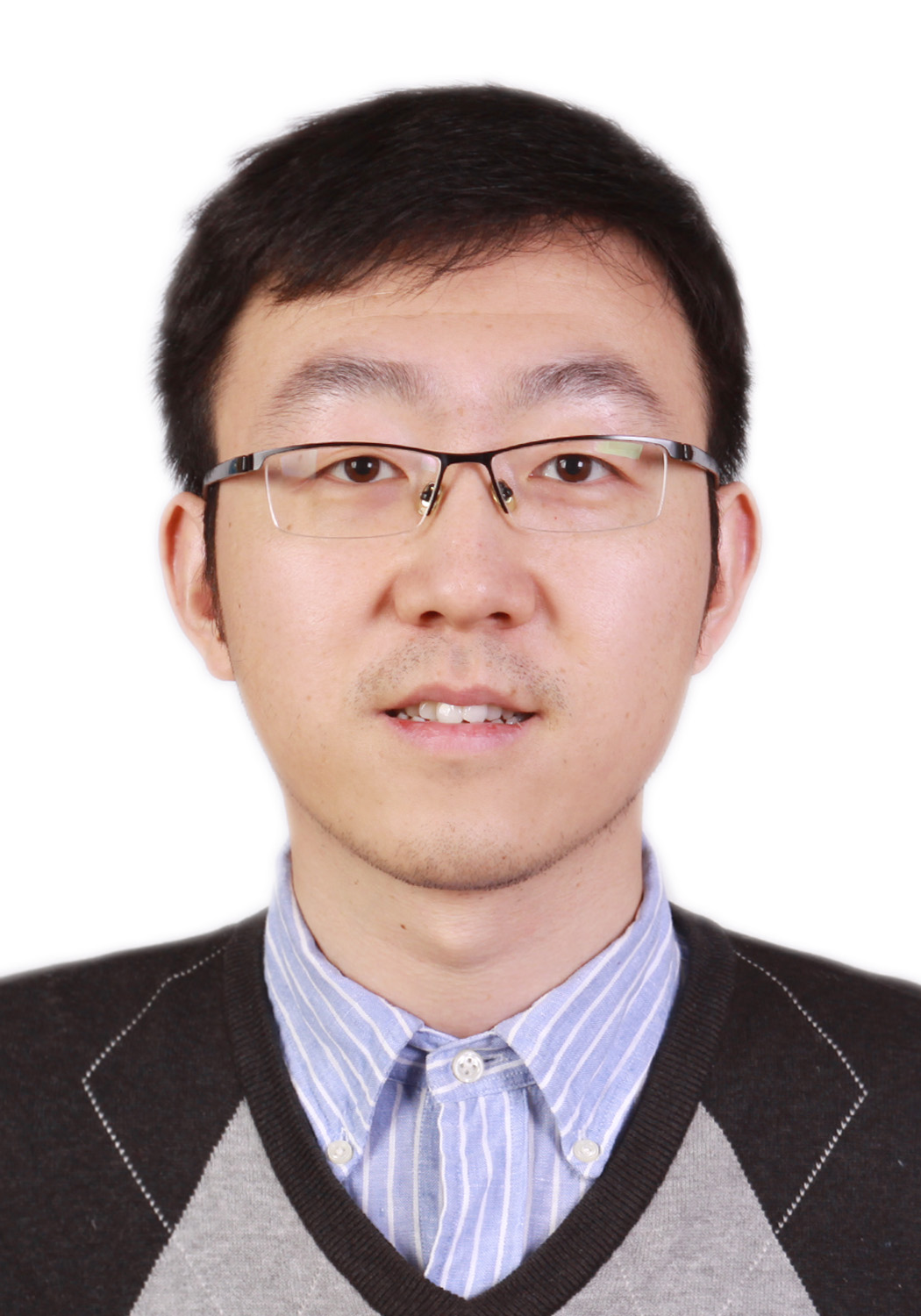}}]{Dong Liu}
received the B.S. and Ph.D. degrees in electrical engineering from the University of Science and Technology of China (USTC), Hefei, China, in 2004 and 2009, respectively.

He was a Member of Research Staff with Nokia Research Center, Beijing, China, from 2009 to 2012. He joined USTC in 2012 and became a Professor in 2020.
His research interests include image and video processing, coding, analysis, and data mining. He has authored or co-authored more than 100 papers in international journals and conferences. He has more than 20 granted patents. He has several technique proposals adopted by standardization groups. He received the 2009 IEEE TCSVT Best Paper Award and the VCIP 2016 Best 10\% Paper Award. He and his students were winners of several technical challenges held in ICCV 2019, ACM MM 2019, ACM MM 2018, ECCV 2018, CVPR 2018, and ICME 2016. He is a Senior Member of CCF and CSIG, an elected member of MSA-TC of IEEE CAS Society, and an elected member of Multimedia TC of CSIG. He serves or had served as the Chair of IEEE 1857.11 Standard Working Subgroup (previously known as FVC-SG), an Associate Editor for Frontiers in Signal Processing, an Organizing Committee Member for VCIP 2022, ICMR 2021, ICME 2021, ICME 2019.
\end{IEEEbiography}

\begin{IEEEbiography}[{\includegraphics[width=1in,height=1.25in,clip,keepaspectratio]{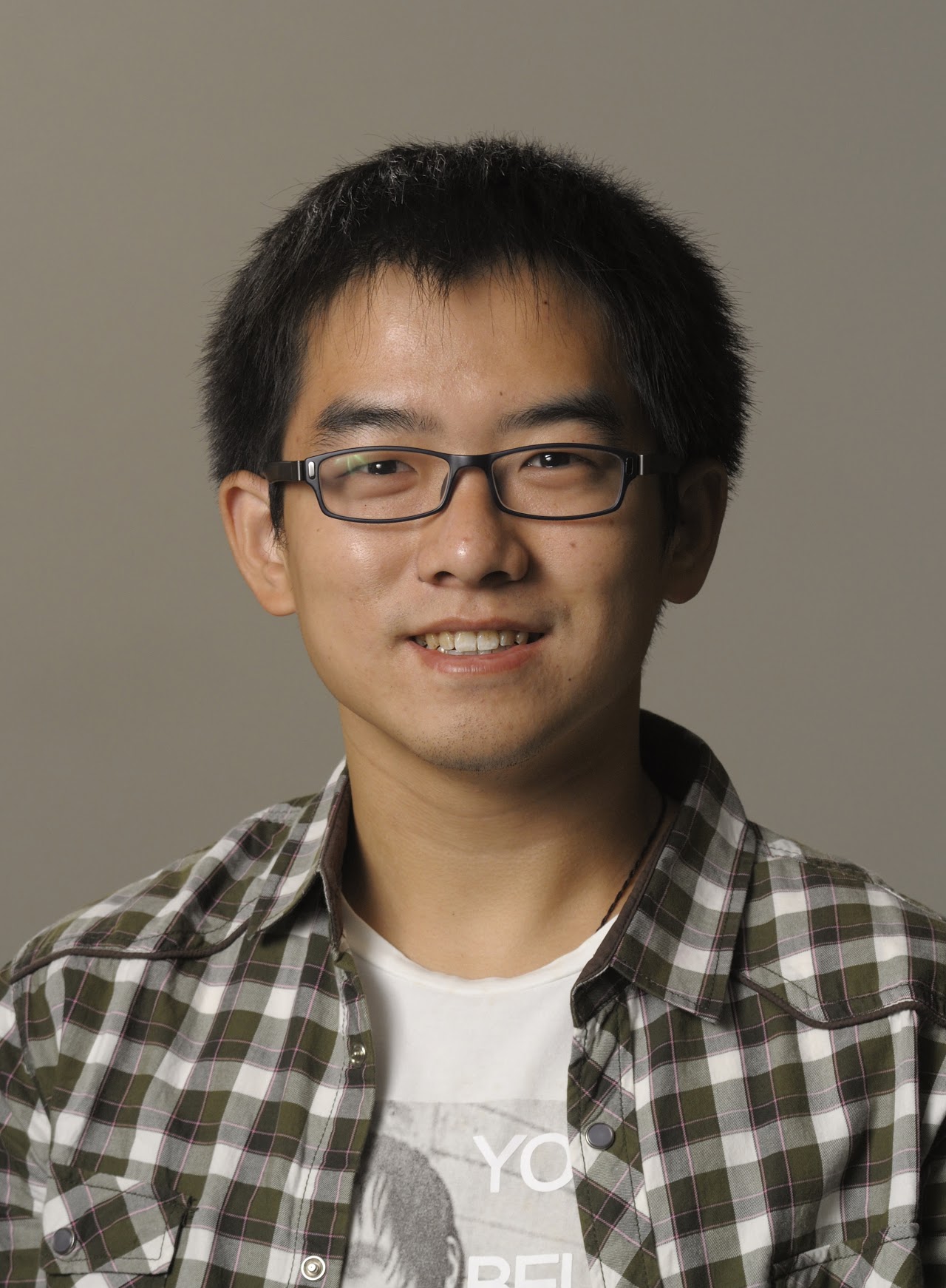}}]{Zhangyang Wang}
is currently an Assistant Professor of Electrical and Computer Engineering at The University of Texas at Austin. He was an  Assistant Professor at Texas A\&M University, from 2017 to 2020. He received his Ph.D. degree from University of Illinois at Urbana–Champaign, under the supervision of Prof.Thomas Huang. Prof. Wang has broad research interests in machine learning, computer vision, optimization, and their interdisciplinary applications. Most recently, he studies automated machine learning (AutoML), learning to optimize (L2O), robust learning, efficient learning, and graph neural networks. 
\end{IEEEbiography}
\end{document}